\documentclass{article}
\usepackage{ijcai24}

\usepackage{times}
\usepackage{soul}
\usepackage{url}
\usepackage[hidelinks]{hyperref}
\usepackage[utf8]{inputenc}
\usepackage[small]{caption}
\usepackage{graphicx}
\usepackage{amsmath}
\usepackage{amsthm}
\usepackage{booktabs}
\usepackage{algorithm}
\usepackage{algorithmic}
\usepackage[switch]{lineno}

\usepackage{multirow} 
\usepackage{diagbox}
\usepackage{amsfonts,amssymb}
\usepackage{bbm}

\usepackage{xcolor}

\title{Denoising-Aware Contrastive Learning for Noisy Time Series}

\author{
Shuang Zhou$^1$
\and
Daochen Zha$^2$\and
Xiao Shen$^3$\and
Xiao Huang$^{1}$\footnote{Corresponding author}\and
Rui Zhang$^4$\And
Fu-Lai Chung$^1$
\affiliations
$^1$Department of Computing, The Hong Kong Polytechnic University\\ $^2$Department of Computer Science, Rice University\\
$^3$School of Computer Science and Technology, Hainan University\\ 
$^4$Department of Surgery, University of Minnesota 
\emails
shuang.zhou@connect.polyu.hk, daochen.zha@rice.edu, shenxiaocam@163.com,\\
zhan1386@umn.edu, \{xiaohuang, cskchung\}@comp.polyu.edu.hk
}

\begin{document}

\maketitle


\begin{abstract}
Time series self-supervised learning (SSL) aims to exploit unlabeled data for pre-training to mitigate the reliance on labels. Despite the great success in recent years, there is limited discussion on the potential noise in the
time series, which can severely impair the performance of existing SSL methods.
To mitigate the noise, the de facto strategy is to apply conventional denoising methods before model training. However, this pre-processing approach may not fully eliminate the effect of noise in SSL for two reasons:
(\textit{i}) the diverse types of noise in time series make it difficult to automatically determine suitable denoising methods;
(\textit{ii}) noise can be amplified after mapping raw data into latent space. 
In this paper, we propose denoising-aware contrastive learning (DECL), which uses contrastive learning objectives to mitigate the noise in the representation and automatically selects suitable denoising methods for every sample.
Extensive experiments on various datasets verify the effectiveness of our method. The code is open-sourced~\footnote{https://github.com/betterzhou/DECL}.

\end{abstract}

\section{Introduction}
Time series learning has attached great importance in various real-world applications~\cite{ismail2019deep}, such as heart failure diagnosis and fault detection in the industry.
Given the abundance of unlabeled time series data~\cite{meng2023unsupervised}, there has been a surge in attention towards time series self-supervised learning (SSL) that extracts informative representations from unlabelled time series data for downstream tasks~\cite{ma2023survey}.
Many time series SSL methods have been proposed in recent years~\cite{zhang2023self_SSL}, including contrastive learning-based~\cite{tonekaboni2021unsupervised}, generative-based~\cite{chowdhury2022tarnet}, and adversarial-based~~\cite{luo2019e2gan} approaches.

\begin{figure}[t]
\begin{center}
\begin{tabular}{c}
\includegraphics[width = 0.95\linewidth]{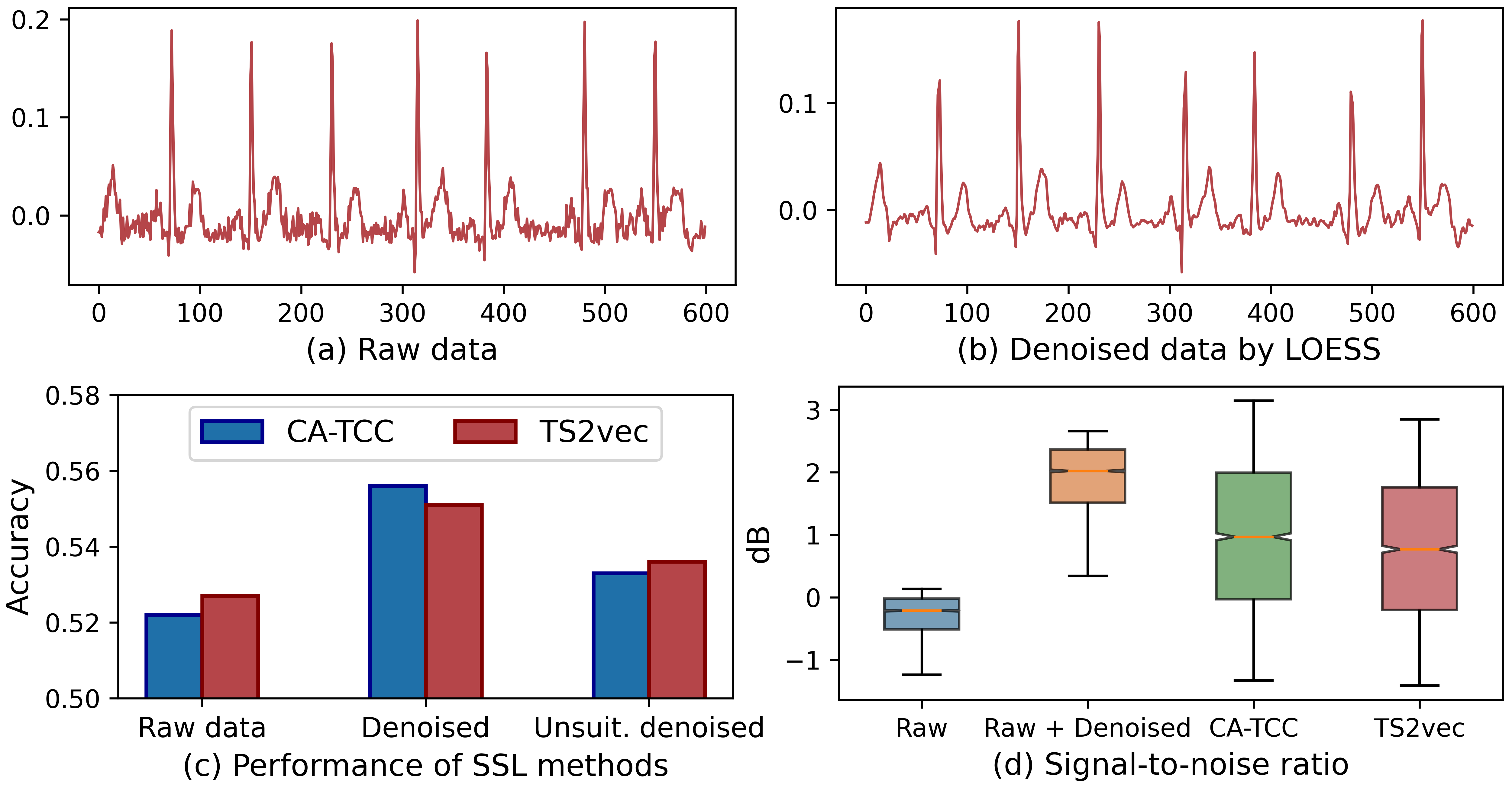}
\end{tabular}
\end{center}
\caption{A motivating analysis (more details provided in Appendix). (a-c) show that the SSL methods achieve higher performance after pre-processing the noisy time series in the PTB-XL dataset with a suitable denoising method LOESS, while the performance improvement is not obvious when processed by unsuitable methods like median filter. (d) suggests SSL methods tend to amplify the noise in the representation.}
\label{Fig1_pilot_noise}
\vspace{-0.3cm}
\end{figure}

Despite the encouraging progress made in time series SSL, the existing research often assumes that the given time series is clean, with limited discussion on the potential noise in the time series. Unfortunately, many time series (e.g., bio-signals collected from sensors) naturally suffer from noises that can severely change the data characteristics and impair representation learned by SSL algorithms~\cite{zhang2023co}. For instance, some samples in the ECG dataset PTB-XL~\cite{wagner2020ptb} exhibit considerable high-frequency noise, depicted in Fig.~\ref{Fig1_pilot_noise}(a). Directly applying the existing time series SSL methods CA-TCC~\cite{eldele2023self} and TS2vec~\cite{yue2022ts2vec} on this dataset for the classification task yields poor accuracy, as illustrated on the left in Fig.\ref{Fig1_pilot_noise}(c). In contrast, by employing an appropriate denoising method like LOESS~\cite{burguera2018fast}, as illustrated in Fig.\ref{Fig1_pilot_noise}(b), the accuracy experiences a significant improvement, as shown in the middle of Fig.\ref{Fig1_pilot_noise}(c).
Motivated by this, we study the following research question: \emph{How can we effectively denoise noisy time series for SSL to learn better representations?}

\setlength{\abovecaptionskip}{-0.04cm}

The de facto strategy to handle noises in time series is to apply conventional denoising methods (e.g., LOESS mentioned above) and then perform model learning~\cite{lai2023practical}. However, we argue that this pre-processing approach cannot fully eliminate the effect of noise in time series.
Firstly, the diverse types of noise in time series make it challenging to select the most suitable denoising methods~\cite {zhang2021eegdenoisenet}.
Many real-world datasets, e.g., ECG data, may contain thousands of samples, where each sample involves different noises~\cite{he2015optimal}, such as high-frequency noise, baseline wandering, and muscle artifacts~\cite{zhang2021heartbeats}.
There is often no denoising method that can universally handle all types of noise~\cite{robbins2020sensitive,zheng202012}, making it difficult to select suitable denoising methods in real-world applications. For example, on the right-hand side of Fig.\ref{Fig1_pilot_noise}(c), the accuracy drops when we apply an unsuitable denoising method median filter.
Secondly, noise can be amplified after mapping the raw data into latent space.
To illustrate this, we compare the signal-to-noise ratio (SNR)~\cite{chawla2011pca} of the raw time series, the denoised time series, and the representations learned upon the denoised time series in Fig.~\ref{Fig1_pilot_noise}(d). The results show that (\textit{i}) the SNR value of the denoised data is improved, meaning that the noise is mitigated in the raw data; (\textit{ii}) however, if further mapping the denoised data into representations, the SNR value drops, which suggests that the representation learning process could amplify the noise. 
That is, even though the denoising methods can alleviate noise in the raw time series, the noise could ``come back'' in the representation space and still hamper SSL.
Hence, how to mitigate the noise in time series SSL remains an open challenge.

In this paper, we propose DEnoising-aware Contrastive Learning (DECL), an end-to-end framework that can leverage any conventional denoising methods to guide noise mitigation in representations.
Specifically, DECL involves two novel designs. Firstly, building upon an auto-regressive encoder, we propose a novel denoiser-driven contrastive learning objective to mitigate the noise.
The key idea is to construct positive samples through the application of existing denoisers on raw time series, and concurrently generating negative samples by introducing noise into the same time series. Subsequently, through optimization using a contrastive learning objective, we guide the representations toward positive samples and distance them from negative samples, thereby reducing the noise. Secondly, we introduce an automatic denoiser selection strategy to learn to select the most suitable denoisers for each sample.
Our motivation is that, in auto-regressive learning, noisy data usually tend to have large reconstruction errors, and vice versa.
As a result, we can use the reconstruction error as a proxy for how suitable a denoiser is to the sample. We further incorporate this denoiser selection strategy into the proposed denoiser-driven contrastive learning and optimize them jointly.
We summarize our contribution below.

\begin{itemize}
\item \textbf{Problem:} Motivated by the observation that removing noise can boost the performance of SSL, we formulate the problem of self-supervised learning on noisy time series.

\item \textbf{Algorithm:} We propose a denoising-aware contrastive learning method that automatically selects suitable denoising methods for each sample to guide mitigating data noise in representation learning.

\item \textbf{Experimental Findings:} Extensive experiments show the effectiveness of our method. We also verify that DECL is robust with varying degrees of noise and the learned representations have less noise. 

\end{itemize}

\begin{figure*}
\begin{center}
\includegraphics[width = 0.9\linewidth]{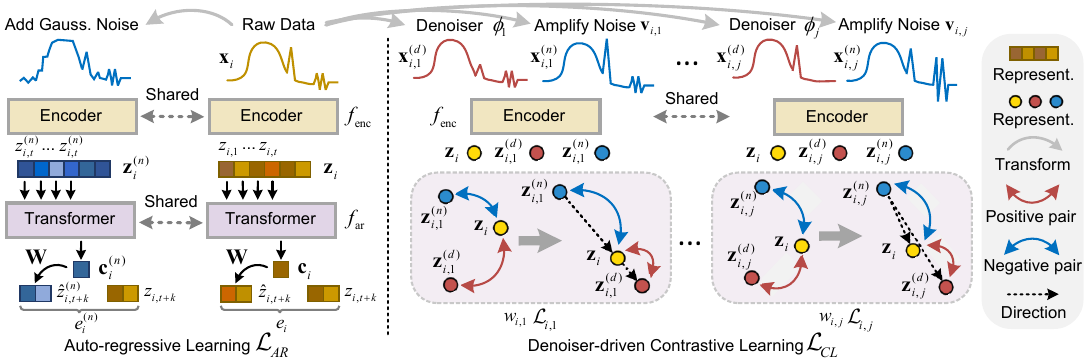}
\end{center}
\caption{Overview of the method DECL. It involves (\textit{i}) auto-regressive learning, which maps raw data into latent space and exploits the representations for SSL; (\textit{ii}) denoiser-driven contrastive learning, which leverages denoising method $\phi_j$ to build positive sample $\mathbf{z}_{i, j}^{(d)}$, amplifies the corresponding noise to build negative sample $\mathbf{z}_{i, j}^{(n)}$, and mitigates noise in representation learning; (\textit{iii}) automatic denoiser selection, which injects Gaussian noise to data to avoid overfitting and determines suitable denoising methods for the contrastive learning.}
\label{Fig2_SSL_framework}
\end{figure*}

\section{Problem Statement}
\noindent\textbf{Self-Supervised Learning on Noisy Time Series.}
Given a set of time series $\mathcal{D} = \left\{\boldsymbol{x}_1, \boldsymbol{x}_2, \ldots, \boldsymbol{x}_N\right\}$ of $N$ instances with a certain amount of noise $\mathcal{S}$, 
the goal is learning a nonlinear mapping function $\mathcal{F}$ in a self-supervised manner that maps each time series $\boldsymbol{x}_i$ to a representation $\mathbf{z}_i$ to best describe itself.
Specifically, for a time series sample $\boldsymbol{x}_i =\left[x_{i, 1}, x_{i, 2}, \ldots, x_{i, T}\right] \in \mathbb{R}^{T \times d}$ with $T$ timestamps and $d$ feature dimensions, it involves noise $\mathbf{s}_i =\left[s_{i, 1}, s_{i, 2}, \ldots,  s_{i, T}\right] \in \mathbb{R}^{T \times d}$ and the denoised data is present as $\boldsymbol{v}_i =\left[(x_{i, 1}-s_{i, 1}), (x_{i, 2} - s_{i, 2}), \ldots,  (x_{i, T} - s_{i, T})\right]$;
the mapping function $\mathcal{F}$ aims to learn a representation $\mathbf{z}_i =\left[z_{i, 1}, z_{i, 2}, \ldots,  z_{i, C}\right] \in \mathbb{R}^{C \times r}$, where ${z}_{i, t} \in \mathbb{R}^r$ is representation at timestamp $t$ with $r$ dimensions.

\section{Methodology}
In this section, we present the proposed DEnoising-aware Contrastive Learning (DECL), as shown in Fig.~\ref{Fig2_SSL_framework}.
It consists of three components: 
(\textit{i}) auto-regressive learning, which is for generating informative representations in latent space; 
(\textit{ii}) denoiser-driven contrastive learning, which exploits denoising methods to guide mitigating noise in representation learning;
(\textit{iii}) automatic denoiser selection, which selects suitable denoising methods for every sample in learning.

\subsection{Auto-regressive Learning}
The purpose is to map raw data into latent space via an encoder and exploit the obtained representations for self-supervised learning.
Specifically, it involves an encoder $f_{\text{enc}}$, which is a $3$-block convolutional architecture, and an auto-regressive (AR) module $f_{\text{ar}}$. 
For an input $\boldsymbol{x}_i$, the encoder maps it to a high-dimensional latent representation $\mathbf{z}_i=f_{\text {enc}}(\boldsymbol{x}_i)$, where $\mathbf{z}_i \in \mathbb{R}^{C \times r}$.
Then, for the representation $\mathbf{z}_i$, the AR module summarizes all $z_{i, j\leq t} = \{z_{i, 1}, z_{i, 2}, ..., z_{i, t}\}$ into a context vector $\mathbf{c}_{i}=f_{ar}\left(z_{i, j\leq t}\right), \mathbf{c}_{i} \in \mathbb{R}^h$, where $h$ is the hidden dimension of $f_{a r}$. 
The context vector $\mathbf{c}_i$ is used to predict the future timesteps from $z_{i, t+1}$ until $z_{i, t+k}(1 \leq k < C)$, $z_{i, t+k} \in \mathbb{R}^r$, where $k$ is the number of predicted timesteps and $r$ is the number of output channels in $f_{\text {enc}}$. 
Here, we use Transformer~\cite{vaswani2017attention} as the $f_{\text{ar}}$, which is comprised of successive blocks of multi-headed attention followed by an MLP block. 
We stack $L$ identical layers to generate the prediction. 
To enable using $\mathbf{c}_i$ to predict the timesteps from $z_{i, t+1}$ until $z_{i, t+k}$, we adopt a linear layer parameterized by $\mathbf{W} \in \mathbb{R}^{h \times r}$ to map $\mathbf{c}_i$ back into the same dimension as $z_i$.
Finally, we obtain the predicted timesteps $\hat{z}_{i, t+1}$ until $\hat{z}_{i, t+k}$.
Accordingly, the reconstruction loss $e_i$ for $\boldsymbol{x}_i$ can be computed by the mean square error between $z_{i, t+j}$ and $\hat{z}_{i, t+j}$:
\begin{equation}
\begin{gathered}
e_i =\frac{1}{k} \sum_{j=1}^k \left(z_{i, t+j}-\hat{z}_{i, t+j}\right)^2,
\end{gathered}
\label{Eq_simple_reconstruction_error}
\end{equation}
where $z_{i,t+j}$ is from $\mathbf{z}_i$ and $\hat{z}_{i,t+j}$ is the prediction. Minimizing the reconstruction loss enables jointly learning $f_{\text{enc}}$ and $f_{\text{ar}}$ for generating informative representations.

\subsection{Denoiser-driven Contrastive Learning}
Considering that noise could be amplified in the latent space, we propose to directly eliminate noise from representations via denoiser-driven contrastive learning.
Our motivation is that conventional denoising methods have been proven effective for noise removal if they are properly used~\cite{zheng202012}; intuitively, the representations of the denoised data suffer from less noise than that of raw data and can be exploited as positive samples to guide learning. 
Vice versa, it is also viable to amplify noise in the raw data, and the representations of the noise-enhanced data are not desired and can act as negative samples.   
Following this, we propose to exploit the noise-reduced and noise-enhanced data in contrastive learning for better representations.

It involves two steps: (1) generating denoised and noise-enhanced counterparts for raw data and obtaining the representations; 
(2) mapping the representations of raw data close to that of the denoised ones (i.e., positive samples) and far away from the noise-enhanced counterparts (i.e., negative samples).
Specifically, we first conduct data augmentation on raw data.
Given a suitable denoising method $\phi_j$ for raw data $\boldsymbol{x}_i$, we can generate the denoised data $\boldsymbol{x}^{(d)}_{i, j}$. 
As for synthesizing noise-enhanced data, the principle is to add more noise w.r.t. the corresponding noise type.
To this end, we compare raw data $\boldsymbol{x}_i$ with the denoised one $\boldsymbol{x}^{(d)}_{i, j}$ to identify what are the ``noises'' and then amplify them. 
In particular, we first obtain the data noise $\boldsymbol{v}_{i, j}$, which has been removed by denoising method $\phi_j$, by using $\boldsymbol{x}_i$ subtracts $\boldsymbol{x}^{(d)}_{i, j}$.
Then, we scale the values of $\boldsymbol{v}_{i, j}$ to further amplify noise and later add it back to $\boldsymbol{x}_i$.
In this way, we obtain the noise-enhanced counterpart $\boldsymbol{x}^{(n)}_{i, j}$ for raw data $\boldsymbol{x}_i$.

After data augmentation, we obtain their representations via the $f_{\text{enc}}$ and perform denoiser-driven contrastive learning.
Given a sample $\boldsymbol{x}_i$ and a denoising method $\phi_j$, we first build a triplet of representations $\mathcal{A}(\mathbf{z}_i, \phi_j)$ as follows:
\begin{equation}
\begin{gathered}
\mathcal{A}(\bold{z}_i, \phi_j) = \{\bold{z}_{i, j}^{(n)}, \bold{z}_i, \bold{z}_{i, j}^{(d)}\} 
\end{gathered}
\label{Eq_triplet_of_emb}
\end{equation}
where $\bold{z}_{i, j}^{(n)}$ and $\bold{z}_{i, j}^{(d)}$ are the representation of $\boldsymbol{x}^{(n)}_{i, j}$ and $\boldsymbol{x}^{(d)}_{i, j}$, and are respectively taken as negative and positive sample in contrastive learning.
Then, by taking representation $\bold{z}_i$ as an anchor, we pull the anchor towards the positive sample while pushing it far away from the negative sample in the latent space. 
Furthermore, considering the triplet of \{$\bold{z}_{i, j}^{(n)}$, $\bold{z}_i$, $\bold{z}_{i, j}^{(d)}$\} consists of the representations with a descending degree of noise, the direction from large noise to trivial noise in the latent space may indicate a better denoising effect. 
Accordingly, we also enforce that the mapping direction of positive and negative samples to the anchor shall be opposite for each triplet. 
Overall, the contrastive learning loss $\mathcal{L}_{i, j}$ for a data $\boldsymbol{x}_i$ and a denoising method $\phi_j$ is shown as:
\begin{equation}
\begin{gathered}
\mathcal{L}_{i, j}= -\log \frac{ \exp \left(\langle \bold{z}_i, \bold{z}_{i, j}^{(d)} \rangle / \tau\right)}{\sum_{\bold{z}_a \in \mathcal{A}(\bold{z}_i, \phi_j)} \mathbbm{1}_{\bold{z}_a \neq \bold{z}_i} \exp \left(\langle \bold{z}_i, \bold{z}_a \rangle / \tau \right)} - \alpha \mathcal{L}_{i, j}^{reg},\\
\mathcal{L}_{i, j}^{reg} = \langle (\bold{z}_i - \bold{z}_{i, j}^{(n)}), (\bold{z}_{i, j}^{(d)} - \bold{z}_{i, j}^{(n)}) \rangle,
\end{gathered}
\label{Eq_contrastive}
\end{equation}
where $\langle\cdot, \cdot\rangle$ denotes cosine similarity, 
$\mathbbm{1}_{\bold{z}_a \neq \bold{z}_i} \in\{0,1\}$ is a binary indicator that equals to 0 when $\bold{z}_a$ denotes $\bold{z}_i$,
$\tau$ is a temporal parameter, 
$\mathcal{L}_{i,j}^{reg}$ is a regularization term that maps representation $\bold{z}_i$ towards a noise-free direction.

\subsection{Automatic Denoiser Selection}

Notably, the above contrastive learning requires a suitable denoising method $\phi_j$ for sample $\boldsymbol{x}_i$. However, it is nontrivial to fulfill this requirement.
Although many conventional denoising methods have been proven effective for noise removal, they may not well handle noise for a given time series $\boldsymbol{x}_i$ when the noise type does not match. 
To address this issue, we propose to collect a set of commonly used denoising methods $\mathcal{M} = \left\{\phi_1, \phi_2, \ldots, \phi_m\right\}$ (shown in Appendix Table~2) in related works and automatically select suitable ones from them.

Here, we propose to exploit the reconstruction error from $f_{\text {ar}}$ to determine suitable methods in $\mathcal{M}$.
Our inspiration is that noisy data usually cause relatively large reconstruction errors.
Following this, the data processed by suitable denoising methods would have small reconstruction errors, while the unsuitable methods would cause large errors. In this way, suitable methods in $\mathcal{M}$ can be automatically determined for every sample $\boldsymbol{x}_i$.
However, directly using the above reconstruction error (Eq.\eqref{Eq_simple_reconstruction_error}) as the learning objective may not reach the goal. 
This is because the AR module may overfit the noisy data, thus rendering raw data to have smaller reconstruction errors than the denoised ones. 
To tackle the issue, we propose another regularization term that encourages $f_{ar}$ to learn global patterns from the noisy time series and avoid overfitting.
The key idea is to augment data with more noise, feed their representations to $f_{ar}$, and enforce reconstructing the raw data that are of less noise.
Specifically, we add Gaussian noise to raw data for data augmentation and map it into the latent space to get the representation $\mathbf{z}^{(n)}_{i}$ as follows: 
\begin{equation}
\begin{gathered}
\mathbf{z}^{(n)}_{i}=\left[z^{(n)}_{i,1}, z^{(n)}_{i,2}, \ldots,  z^{(n)}_{i, C}\right].
\end{gathered}
\label{Eq_z_aug_emb}
\end{equation}
Similarly, we feed the $z^{(n)}_{i, j\leq t}$ into $f_{\text{ar}}$ to obtain $\mathbf{c}_i^{(n)}$ and the prediction $\hat{z}^{(n)}_{i, t+j}$.
Formally, the overall learning objective for auto-regressive learning is:
\begin{equation}
\begin{gathered}
\mathcal{L}_{AR}=\frac{1}{N} \sum_{i=1}^N \left( e_i + e_i^{(n)} \right),
e_i^{(n)} =\frac{1}{k} \sum_{j=1}^k \left(z_{i, t+j}-\hat{z}_{i, t+j}^{(n)}\right)^2,
\end{gathered}
\label{Eq_overall}
\end{equation}
where $N$ is the number of training samples.
During optimization, $\mathcal{L}_{AR}$ encourages capturing global patterns from the noisy time series for prediction, which avoids overfitting the raw data.
Thereby, the samples processed by suitable denoising methods would eventually own smaller reconstruction errors than the raw data, and it is feasible to use reconstruction errors for the automatic selection.

Considering that a data sample may contain multiple types of noises and require different denoising methods, we leverage all the suitable methods in $\mathcal{M}$ by assigning them large weights while setting small weights for the unsuitable ones.
Specifically, we first feed the processed sample $\boldsymbol{x}_{i,j}^{(d)}$ by a denoising method $\phi_j$ to the optimized $f_{\text{ar}}$ and obtain the reconstruction error $e_{i,j}$ based on the Eq.~\eqref{Eq_simple_reconstruction_error}. 
Then, we compute the weight value $w_{i, j}$ for $\phi_j$ with a softmax function as below:
\begin{equation}
\begin{gathered}
w_{i,j} =\frac{ e_{i,j}^{-1}}{\sum_{j=1}^m e_{i,m}^{-1} },
\end{gathered}
\label{Eq_compute_weights}
\end{equation}
where the $m$ is the number of methods in $\mathcal{M}$.
Hence, the contrastive learning objective $\mathcal{L}_{CL}$ is a weighted combination of $\mathcal{L}_{\textit{i, j}}$ for the denoising methods in $\mathcal{M}$, i.e.,
\begin{equation}
\begin{gathered}
\mathcal{L}_{CL}= \sum_{i=1}^{N} \sum_{j=1}^{m} w_{i, j} \mathcal{L}_{\textit{i, j}},
\end{gathered}
\label{Eq_weighted_CL}
\end{equation}
where $N$ is the number of the training samples, $w_{i, j}$ is the weight score for method $\phi_j$.
By minimizing $\mathcal{L}_{CL}$, our method encourages mapping the representations of raw data toward a noise-free direction, thus mitigating the data noise in latent space.
Finally, we combine auto-regressive learning and contrastive learning for joint optimization:
\begin{equation}
\begin{gathered}
\mathcal{L} = \gamma \mathcal{L}_{AR} + \mathcal{L}_{CL},
\end{gathered}
\label{Eq_Loss_final}
\end{equation}
where $\gamma$ is a weight value to balance the two terms.

\section{Experiments}

We perform empirical evaluations to answer the following research questions:
\textbf{RQ1:} How effective is DECL for unsupervised representation learning?
\textbf{RQ2:} Is the method effective with fine-tuning?
\textbf{RQ3:} What are the effects of each component? 
\textbf{RQ4:} Is it robust with varied degrees of noise?
\textbf{RQ5:} How sensitive is it to the hyper-parameters?
\textbf{RQ6:} How does the method work in practice?

\begin{table}[]
\centering
\caption{Statistics of the noisy time series datasets.}
\resizebox{0.95\linewidth}{!}{
\begin{tabular}{ccccccc}
\toprule
\textbf{Datasets}       & \textbf{\# Train} & \textbf{\# Valid} & \textbf{\# Test} & \textbf{Length} & \textbf{\# Channel} & \textbf{\# Class} \\
\midrule
\textbf{SleepEDF}       & 16,923            & 8,462             & 16,923           & 3,000           & 1                & 5                 \\
\textbf{FaultDiagnosis} & 5,456             & 2,728             & 5,456            & 5,120           & 1                & 3                 \\
\textbf{CPSC18}         & 13,754            & 6,877             & 13,754           & 2,000           & 12               & 9                 \\
\textbf{PTB-XL}         & 6,509             & 3,254             & 6,509            & 2,000           & 12               & 5                 \\
\textbf{Georgia}        & 6,334             & 3,167             & 6,334            & 2,000           & 12               & 6               \\ 
\bottomrule
\end{tabular}
}
\label{Table_dataset_statistic}
\end{table}

\subsection{Dataset}
We employ five noisy time series datasets. 
SleepEDF~\cite{goldberger2000physiobank} is an EEG dataset in which each sample records human brain activity. 
The data in FaultDiagnosis~\cite{lessmeier2016condition} are collected from sensor readings of bearing machine under different working conditions. 
CPSC18~\cite{liu2018open}, PTB-XL~\cite{wagner2020ptb}, and Georgia~\cite{alday2020classification} are ECG datasets wherein each sample reflects heart activity.
The data statistics are shown in Table~\ref{Table_dataset_statistic}.
See Appendix~\ref{Datasets} for more details.

\begin{table*}[t]
\centering
\caption{Overall performance (\%) comparison on the datasets. \textit{DN} means pre-processed by a suitable denoising method. 
}
\resizebox{0.95\linewidth}{!}{
\begin{tabular}{ccccccccccc}
\toprule
\multirow{2}{*}{\textbf{Methods}} & \multicolumn{2}{c}{\textbf{SleepEDF}} & \multicolumn{2}{c}{\textbf{FaultDiagnosis}} & \multicolumn{2}{c}{\textbf{CPSC18}} & \multicolumn{2}{c}{\textbf{PTB-XL}} & \multicolumn{2}{c}{\textbf{Georgia}} \\
\cmidrule(r){2-3} \cmidrule(r){4-5} \cmidrule(r){6-7} \cmidrule(r){8-9} \cmidrule(r){10-11}
                                  & \textit{Accuracy}         & $\textit{Weighted-F}_{1}$        & \textit{Accuracy}         & $\textit{Weighted-F}_{1}$        & \textit{Accuracy}        & $\textit{Weighted-F}_{1}$       & \textit{Accuracy}        & $\textit{Weighted-F}_{1}$       & \textit{Accuracy}        & $\textit{Weighted-F}_{1}$        \\
\midrule
TF-C                              & 62.07±0.61        & 61.24±0.54        & 76.82±0.72           & 74.69±0.71           & 38.46±0.50       & 35.91±0.64       & 51.21±0.62       & 40.93±0.37       & 48.97±0.34       & 33.43±0.38        \\
TF-C + \textit{DN}                         & 65.49±0.67        & 64.83±0.61        & 78.39±0.64           & 77.54±0.75           & 40.53±0.57       & 37.15±0.52       & 53.19±0.65       & 43.24±0.48       & 51.29±0.45       & 35.18±0.39        \\
TF-C + \textit{Merge}                      & 61.72±0.72        & 60.49±0.67        & 77.51±0.85           & 75.33±0.70           & 39.07±0.68       & 36.03±0.56       & 52.54±0.71       & 41.59±0.47       & 49.58±0.59       & 33.97±0.43        \\
TS2vec                            & 63.68±0.56        & 62.75±0.42        & 77.63±0.49           & 77.15±0.44           & 40.56±0.59       & 37.47±0.47       & 52.71±0.59       & 44.37±0.21       & 51.61±0.43       & 36.42±0.32        \\
TS2vec + \textit{DN}                       & 67.41±0.63        & 67.92±0.47        & 81.15±0.33           & 80.39±0.25           & 42.83±0.63       & 39.08±0.51       & 55.14±0.32       & 46.18±0.16       & 53.27±0.28       & 38.48±0.17        \\
TS2vec + \textit{Merge}                    & 63.07±0.70        & 62.38±0.56        & 79.29±0.56           & 78.24±0.48           & 40.98±0.72       & 37.63±0.69       & 53.48±0.54       & 45.24±0.32       & 52.19±0.41       & 36.93±0.34        \\
CRT                               & 62.25±0.48        & 61.03±0.38        & 75.82±0.58           & 76.07±0.32           & 39.25±0.65       & 36.34±0.34       & 52.32±0.58       & 42.31±0.35       & 49.92±0.27       & 34.51±0.15        \\
CRT + \textit{DN}                          & 65.90±0.39        & 65.16±0.15        & 78.93±0.41           & 78.49±0.26           & 41.92±0.51       & 37.96±0.38       & 54.17±0.36       & 44.92±0.14       & 52.17±0.24       & 36.34±0.21        \\
CRT + \textit{Merge}                       & 61.94±0.53        & 61.21±0.39        & 76.75±0.54           & 77.04±0.37           & 39.83±0.79       & 36.53±0.53       & 52.89±0.42       & 43.12±0.23       & 50.61±0.35       & 34.79±0.26        \\
SimMTM                            & 63.84±0.57        & 62.95±0.43        & 78.39±0.43           & 77.52±0.39           & 40.74±0.56       & 37.72±0.32       & 52.63±0.43       & 44.75±0.39       & 51.34±0.32       & 35.63±0.24        \\
SimMTM + \textit{DN}                       & 68.62±0.41        & 68.19±0.24        & 81.07±0.32           & 80.93±0.08           & 43.21±0.44       & 39.65±0.25       & 55.18±0.37       & 46.39±0.16       & 53.82±0.29       & 38.15±0.23        \\
SimMTM + \textit{Merge}                    & 63.31±0.52        & 62.27±0.36        & 79.64±0.50           & 78.43±0.33           & 41.30±0.62       & 38.14±0.41       & 53.41±0.51       & 45.17±0.40       & 52.19±0.23       & 36.32±0.32        \\
TS-CoT                            & 64.62±0.59        & 63.98±0.28        & 76.32±0.44           & 75.96±0.32           & 39.82±0.78       & 37.43±0.39       & 52.92±0.54       & 43.14±0.35       & 50.97±0.31       & 34.85±0.21        \\
TS-CoT + \textit{DN}                       & 66.21±0.44        & 67.05±0.31        & 79.41±0.37           & 78.65±0.27           & 41.36±0.53       & 38.57±0.25       & 54.28±0.49       & 45.30±0.24       & 52.83±0.25       & 37.09±0.08        \\
TS-CoT + \textit{Merge}                    & 64.29±0.52        & 63.17±0.45        & 77.83±0.46           & 76.42±0.31           & 40.27±0.74       & 37.64±0.46       & 53.32±0.66       & 43.85±0.27       & 51.38±0.46       & 35.41±0.16        \\
CA-TCC                            & 63.91±0.48        & 63.37±0.33        & 78.05±0.43           & 77.39±0.26           & 40.79±0.54       & 38.29±0.49       & 52.27±0.32       & 44.91±0.22       & 52.07±0.27       & 36.26±0.34        \\
CA-TCC + \textit{DN}                       & 68.48±0.27        & 68.11±0.21        & 81.39±0.29           & 80.74±0.15           & 43.67±0.47       & 39.87±0.24       & 55.54±0.25       & 46.12±0.13       & 53.69±0.24       & 38.37±0.29        \\
CA-TCC + \textit{Merge}                    & 63.67±0.43        & 63.09±0.37        & 79.81±0.35           & 78.26±0.28           & 41.18±0.61       & 38.95±0.43       & 53.11±0.38       & 45.33±0.18       & 52.43±0.33       & 36.84±0.37        \\
\midrule
\textbf{Only AR-}                          & 62.35±0.51        & 61.77±0.42        & 76.95±0.41           & 75.81±0.32           & 38.64±0.56       & 36.26±0.32       & 51.80±0.53       & 41.39±0.35       & 50.42±0.28       & 34.29±0.35        \\
\textbf{Only AR}                           & 62.89±0.56        & 62.31±0.36        & 77.46±0.52           & 76.73±0.39           & 39.22±0.48       & 36.41±0.21       & 52.27±0.44       & 41.94±0.29       & 51.16±0.17       & 35.03±0.29        \\
\textbf{CL}                                & 67.63±0.39        & 67.34±0.19        & 80.84±0.36           & 80.21±0.11           & 42.65±0.52       & 38.93±0.37       & 54.89±0.52       & 45.72±0.24       & 52.98±0.25       & 37.62±0.27        \\
\textbf{DECL-}                             & 68.35±0.42        & 67.87±0.34        & 81.16±0.47           & 80.58±0.34           & 43.12±0.33       & 39.16±0.26       & 55.06±0.31       & 45.93±0.16       & 53.29±0.20       & 38.01±0.14        \\
\textbf{DECL*}                             & 70.19±0.48        & 69.52±0.30        & 81.53±0.33           & 81.26±0.29           & 43.96±0.49       & 39.84±0.34       & 56.30±0.37       & 48.24±0.18       & 54.65±0.12       & 39.48±0.23        \\
\textbf{DECL  (Ours)}             & \textbf{71.74±0.43}        & \textbf{70.96±0.25}        & \textbf{82.78±0.38}           & \textbf{82.17±0.23}           & \textbf{45.01±0.32}       & \textbf{41.65±0.39}       & \textbf{57.41±0.36}       & \textbf{49.32±0.27}       & \textbf{55.37±0.26}       & \textbf{40.35±0.21}       \\
\bottomrule
\end{tabular}
}
\label{Table_1_overall}
\end{table*}

\subsection{Experimental Settings}
$\\$
\noindent\textbf{Comparative Methods.}
We compare our method with representative SSL methods, including contrastive learning-based, such as TF-C~\cite{zhang2022self}, TS2vec~\cite{yue2022ts2vec}, TS-CoT~\cite{zhang2023co}, and CA-TCC~\cite{eldele2023self}, as well as generative-based methods, e.g., CRT~\cite{zhang2023self} and SimMTM~\cite{dong2023simmtm}. We provide detailed descriptions in Appendix~\ref{SSL_Methods}.

\noindent\textbf{Evaluation Metrics.}
We follow previous works~\cite{eldele2023self}\cite{yue2022ts2vec} and adopt $\textit{Accuracy}$ and $\textit{Weighted-F}_{1}$ scores for classification performance evaluation.

\noindent\textbf{Implementation Details.}
We split the data into $40$\%, $20$\%, and $40$\% for training, validation, and test set.
We set the learning epochs as $100$ and adopt a batch size of $128$ for both pre-training and downstream tasks, as we notice the training loss does not further decrease. 
In the transformer, we set $L$ as $4$, the number of heads as $4$, and the hidden dimension size as $100$. The details of the encoder and AR module can be referred to in Appendix~\ref{appendx_exp_DECL_details}.
As for the hyper-parameters, we set $k$ as $30$\% of the total timestamps, assign $\alpha$ as $0.5$, and set $\gamma$ as $0.1$ for all the datasets. 
The method is optimized with Adam optimizer; we set the learning rate as $1e$-$4$ and the weight decay as $5e$-$4$. 
We collect the commonly used denoising methods from related papers for different types of time series and apply the reported hyper-parameters or the default ones for noise removal (see the details in Appendix Table~\ref{Table_denoise_summary}).
For the baselines, we select the hyper-parameters with the best performance on the validation set for the downstream task. 
For a fair comparison, we pre-process the raw data for all the baselines by (\textit{i}) applying each denoising method in $\mathcal{M}$ for the dataset and reporting the one with the best performance (i.e., \textit{DN}), and (\textit{ii}) combining all the denoising methods for noise removal (i.e., \textit{Merge}). 
We run experiments $10$ times and report the averaged results.

\subsection{Linear Evaluation of Representations (RQ1)}
We first pre-train SSL methods with unlabeled data for representation learning, then use a portion of labeled data (i.e., $10$\% and $30$\%) to evaluate the effectiveness of the learned representations. 
Following the standard linear evaluation scheme~\cite{eldele2023self}, we fix the parameters of the self-supervised pre-trained model and regard it as an encoder, and then train a linear classifier (single fully connected layer) on top of the encoder. 
The results with $10$\% and $30$\% training labels are shown in Table~\ref{Table_1_overall} and Appendix Table~\ref{Table_1_overall_appedx}.
We find that existing SSL methods achieve sub-optimal performance on noisy data and the performances get boosted after applying a suitable denoising method for noise mitigation. 
Besides, directly combining all the denoising methods for pre-processing causes unsatisfactory performance. 
This can be explained by that certain denoising methods may induce extra noise into data when the noise type does not match. 
Additionally, DECL achieves superior performance than other SSL methods.
Specifically, the performance boost over the best baseline on the CPSC18 dataset is about 3\%. 
This is because our method exploits suitable denoising methods to mitigate noise and guide representation learning.

\subsection{Fine-tuning Performance Evaluation (RQ2)}
\begin{figure}[t]
\begin{center}
\begin{tabular}{c}
\includegraphics[width = 0.98\linewidth]{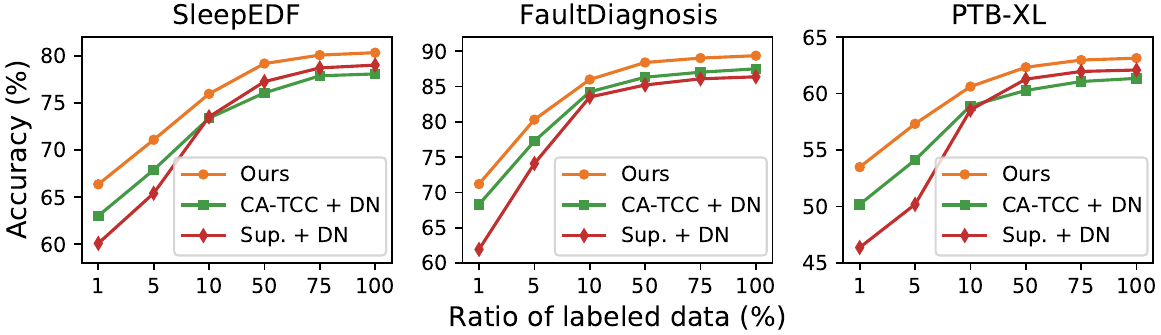}
\end{tabular}
\end{center}
\caption{Performance comparison for semi-supervised representation learning with different percentages of labeled data.}
\label{Fig3_semi_efficiency}
\vspace{-0.2cm}
\end{figure}

To simulate real-world scenarios where a few labeled data are accessible, we fine-tune our pre-trained model with the labels and investigate its effectiveness. There are two setups. $\\$
\noindent\textbf{Fine-tuning on the Source Dataset.} 
Following previous works~\cite{lan2022intra}, we pre-train the model with unlabeled data and then fine-tune it with different amounts of the training labels (i.e., $1$\%, $5$\%, $10$\%, $50$\%, $75$\%, and $100$\%) from the same dataset. 
Fig.~\ref{Fig3_semi_efficiency} shows the performance comparison against its supervised learning counterpart and a strong baseline on three datasets.
We observe that supervised learning performs poorly with limited labeled data (e.g., 1\%), while the fine-tuned models achieve significantly better performance. 
It verifies that self-supervised pre-training can alleviate the label scarcity issue.
Besides, DECL outperforms the strong baseline when using different ratios of training labels in fine-tuning. In brief, it shows the effectiveness of our method under the fine-tuning mode.

\noindent\textbf{Fine-tuning on New Dataset.} 
To evaluate the generalizability of the learned representations, we further pre-train the model on one dataset and perform supervised fine-tuning on another dataset. Specifically, we follow the one-to-one evaluation scheme~\cite{zhang2022self} and use a portion of labeled data (10\%) for fine-tuning. 
Table~\ref{Table_cross_evaluation} shows the results under six cross-dataset scenarios on the ECG data. 
Similarly, we find that combining all the denoising methods would render unsatisfactory performance for the baselines than that of applying a suitable one.
Besides, our fine-tuned model consistently outperforms the strong baselines. 
Furthermore, pre-training on a comprehensive dataset (e.g., CPSC18) usually promises better performance on the new dataset, which is consistent with the findings from related papers~\cite{yang2023biot}.
Overall, the results verify that DECL can alleviate the impact of noise on learning informative representations and generalize well on cross-dataset scenarios.

\begin{table}[t]
\centering
\caption{Performance (accuracy \%) of the transferability evaluation on CPSC18 (C), PTB-XL (P), and Georgia (G) datasets.}
\resizebox{0.95\linewidth}{!}{
\begin{tabular}{ccccccc}
\toprule
\textbf{Methods} & \textbf{P} $\rightarrow$ \textbf{C} & \textbf{G} $\rightarrow$ \textbf{C} & 
\textbf{C} $\rightarrow$ \textbf{P} & 
\textbf{G} $\rightarrow$ \textbf{P} & 
\textbf{C} $\rightarrow$ \textbf{G} & 
\textbf{P} $\rightarrow$ \textbf{G}       \\
\midrule
SimMTM + \textit{DN} & 43.74                & 42.59                     & 55.48                  & 54.59                     & 54.73                      & 52.62                    \\
SimMTM + \textit{Merge} & 41.62                 & 40.81                      & 53.96                  & 52.18                     & 52.80                      & 50.91                    \\
CA-TCC + \textit{DN} & 44.17                 & 42.83                      & 57.43                  & 56.21                     & 54.52                      & 52.47                    \\
CA-TCC + \textit{Merge} & 42.61                 & 40.75                      & 55.69                  & 54.74                     & 53.25                      & 51.16                    \\
DECL (Ours)             & \textbf{46.23}                 & \textbf{44.25}                      & \textbf{60.21}                  & \textbf{59.03}                     & \textbf{56.41}                      & \textbf{55.69}                   \\
\bottomrule
\end{tabular}
}
\label{Table_cross_evaluation}
\end{table}

\subsection{Ablation Study (RQ3)}
We also examine the effect of each component in DECL on its overall performance. 
Specifically, we derive different method variants for comparison:
(1) DECL-, which drops the regularization term in the auto-regressive learning and remains others unchanged;
(2) DECL*, which deletes the direction constraint in the contrastive learning;
(3) CL, which assigns equal weights to the denoising methods and keeps others unchanged;
(4) Only AR, which merely leverages $\mathcal{L}_{AR}$ as the learning objective; 
(5) Only AR-, which exploits $\mathcal{L}_{AR}$ without the regularization term.
The results are shown in Table~\ref{Table_1_overall}.
We find that (\textit{i}) merely using $\mathcal{L}_{AR}$ renders unsatisfactory performance. This is because the noise in latent space cannot be eliminated and hampers representation learning. (\textit{ii}) Our method outperforms CL and DECL-, illustrating that attaching equal importance to the denoising methods would hinder representation learning and the regularization term in $\mathcal{L}_{AR}$ is conducive to the overall performance.
(\textit{iii}) Our method achieves higher performance than DECL*, showing that the direction constraint in contrastive learning helps to eliminate the noise in representation learning.

\begin{figure}
\begin{center}
\begin{tabular}{c}
\includegraphics[width = 0.98\linewidth]{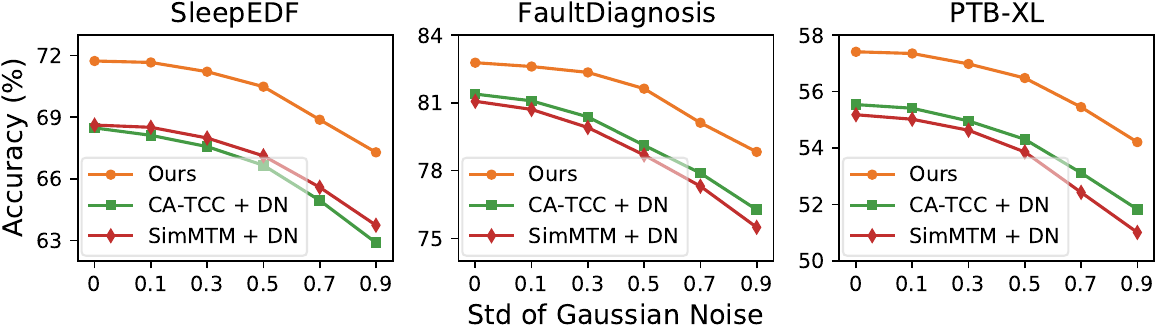}
\end{tabular}
\end{center}
\caption{Unsupervised representation learning performance under varying degrees of data noise.}
\label{Fig5_robustness_noise}
\vspace{-0.3cm}
\end{figure}

\subsection{Robustness Analysis against Data Noise (RQ4)}
We further investigate the robustness of DECL against varying degrees of data noise.
Specifically, we induce Gaussian noise to the raw data (with zero mean and varying standard deviations) and then conduct the linear evaluation for performance comparison. 
The results on three datasets are shown in Fig.~\ref{Fig5_robustness_noise}; see Appendix~\ref{appendx_exp_robustness} for more results.
We have two interesting findings. (\textit{i}) By inducing a larger amount of noise, the performances of the methods gradually drop. This occurs since the adopted denoising method(s) cannot eliminate the noise sufficiently, thereby impairing representation learning.  
(\textit{ii}) Our method consistently outperforms the strong baselines. Because DECL selects suitable denoising methods for every sample and further diminishes noise from representations with contrastive learning.

\begin{figure}[t]
\begin{center}
\begin{tabular}{c}
\includegraphics[width = 0.95\linewidth]{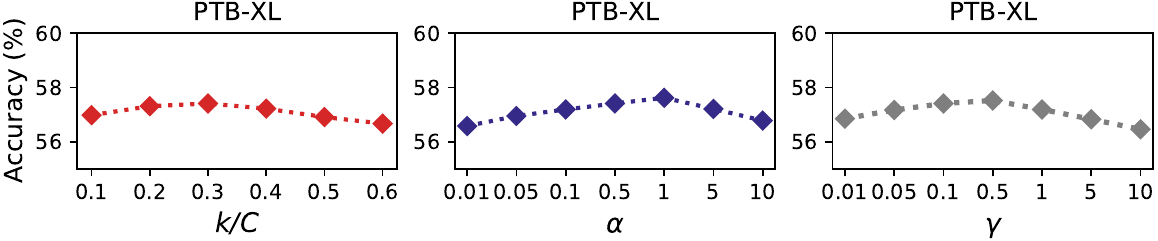}
\end{tabular}
\end{center}
\caption{Hyper-parameter analysis results.}
\label{Fig_hyperparam_PTB}
\vspace{-0.3cm}
\end{figure}

\begin{figure}[t]
\begin{center}
\begin{tabular}{c}
\includegraphics[width = 0.95\linewidth]{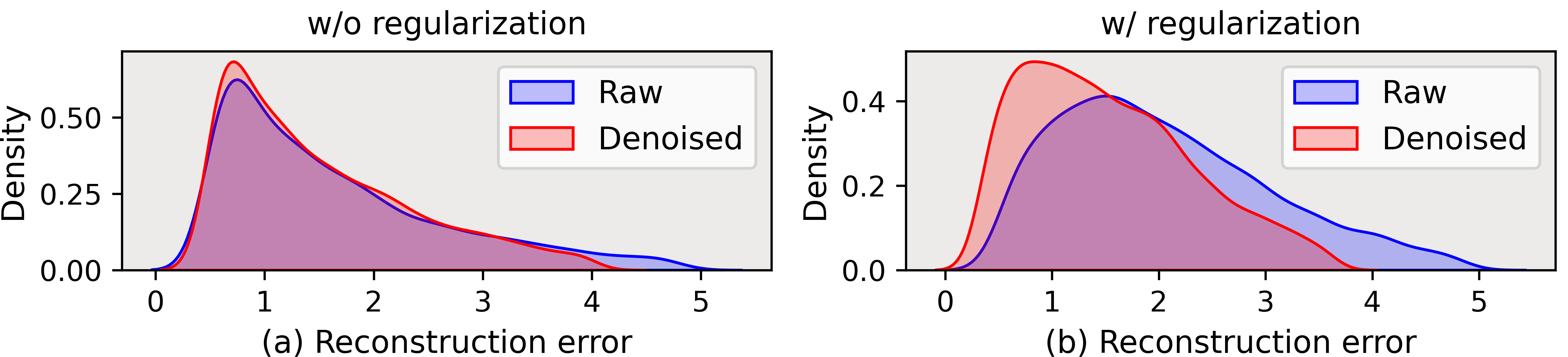}
\end{tabular}
\end{center}
\caption{Comparison of the reconstruction error distribution between raw data and denoised data. After adding regularization, the distribution gap is more distinguishable.}
\label{Fig_reconstruction_error}
\end{figure}

\subsection{Sensitivity Analysis (RQ5)}
We perform sensitivity analysis to study the main hyper-parameters: the number of predicted future timesteps $k$ in Eq.~\eqref{Eq_overall}, the weight $\alpha$ in Eq.~\eqref{Eq_contrastive}, and the weight $\gamma$ in Eq.~\eqref{Eq_Loss_final}.
Specifically, we adopt the same setup as the linear evaluation experiment, present the results of the PTB-XL dataset in Fig.~\ref{Fig_hyperparam_PTB}, and show more results in Appendix~\ref{appendx_exp_hyperparameter}.
We first analyze the impact of $k$, where the \textit{x}-axis denotes the percentage $k/C$ and $C$ is the total number of timesteps. It shows that, as the percentage value increases, the performance first boosts and then declines. Hence, we suggest setting it to the scope of $0.1$-$0.4$ in practice.
Regarding the hyper-parameter $\alpha$ and $\gamma$, when raising its value, the performance first rises and later declines. It occurs because either a small or a large value fails to achieve a balance between the learning objectives. Given this, we recommend setting the value of $\alpha$ to $0.1$-$5$ and the value of $\gamma$ to $0.1$-$1$.

\subsection{Visualization Results (RQ6)}
\noindent\textbf{The Effect of the Regularization Term.} 
Here, we examine whether the proposed regularization term in $\mathcal{L}_{AR}$ (see Eq.~\eqref{Eq_overall}) can alleviate the overfitting issue and enable the reconstruction error as an indicator for choosing suitable denoising methods.
Specifically, we compare the model learned with $\mathcal{L}_{AR}$ (i.e., with regularization) and the counterpart without regularization by visualizing the distribution of reconstruction errors on the PTB-XL dataset. See Appendix~\ref{appendx_exp_reconstruction} for more details.
The results in Fig.~\ref{Fig_reconstruction_error} show that (\textit{i}) without a regularization term, the distributions of reconstruction errors between raw and denoised data become similar, rendering it hard to use the reconstruction error as the indicator.
(\textit{ii}) After adding the regularization term, the distributions of reconstruction errors are more distinguishable.
In brief, it verifies that the proposed regularization term indeed alleviates overfitting and benefits determining suitable denoising methods.

\begin{figure}[t]
\begin{center}
\begin{tabular}{c}
\includegraphics[width = 0.95\linewidth]{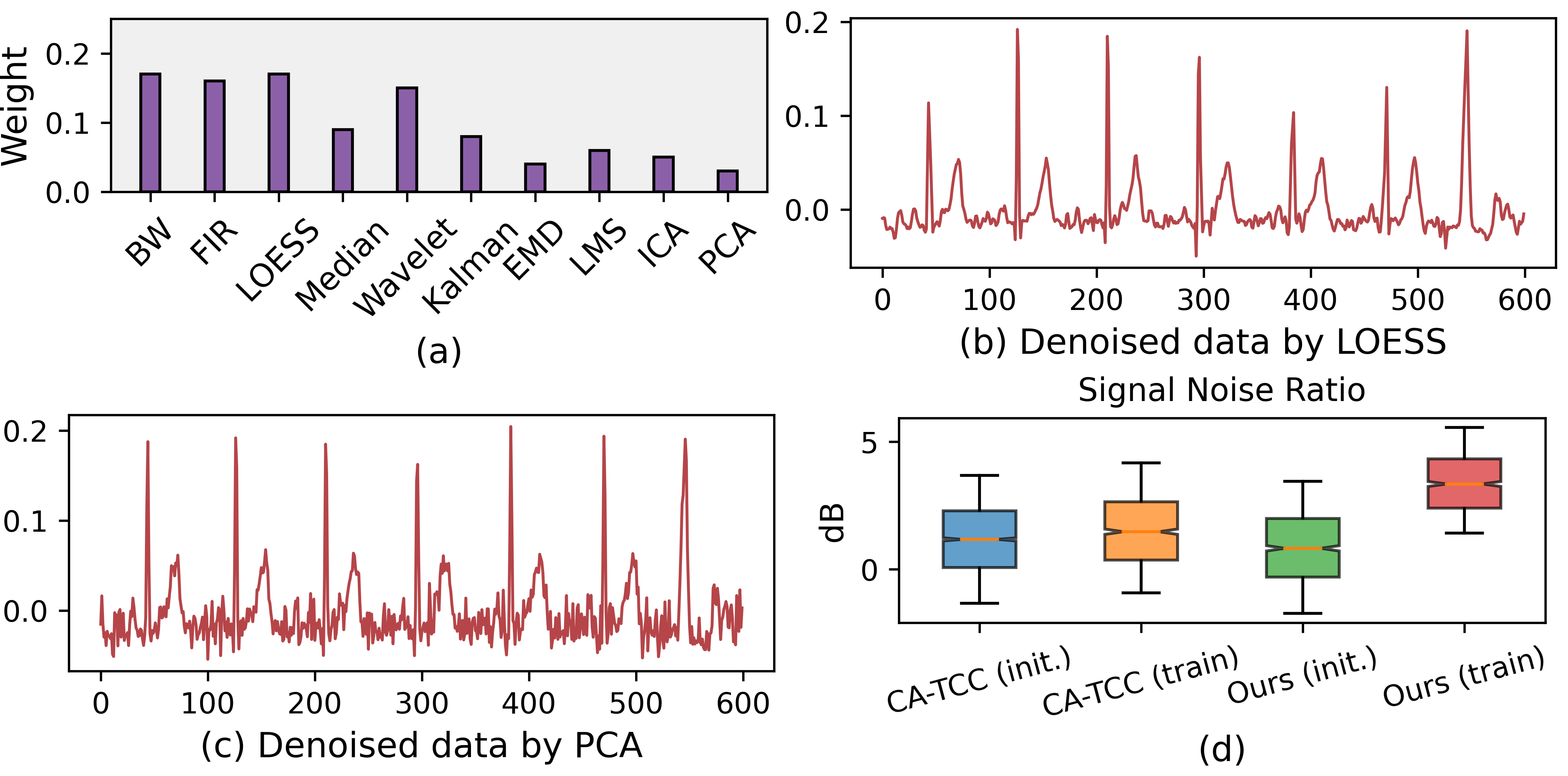}
\end{tabular}
\end{center}
\caption{(a-c) The learned weights for denoising methods and their effects. (d) The comparison of SNR on representation.}
\label{Fig_plot_wave}
\end{figure}

\noindent\textbf{Analysis on the Selected Denoising Methods.} 
We also verify whether DECL can assign more weights to the suitable denoising methods. In detail, we adopt a case study on the PTB-XL dataset to (\textit{i}) visualize the weight values of the denoising methods in $\mathcal{M}$ for a sample with high-frequency noise, and (\textit{ii}) showcase a raw data and the denoised counterparts to examine the denoising effect.
As shown in Fig.~\ref{Fig_plot_wave}(a), the AR module assigns different weights on the set of potentially feasible denoising methods. 
For example, the LOESS method, which is reported to remove high-frequency noise for ECG~\cite{burguera2018fast}, obtains large a weight value; whereas the PCA method, which does not suit this type of noise~\cite{alickovic2015effect}, owns a small value.
We further visualize the denoised data by LOESS in Fig.~\ref{Fig_plot_wave}(b), which demonstrates that noises are well mitigated. However, as shown in Fig.~\ref{Fig_plot_wave}(c), the denoised data by PCA still involve severe noises.
More results are presented in Appendix~\ref{appendx_exp_denoising_methods}.
The above analyses prove that our method can automatically select suitable denoising methods for the noisy time series.

\noindent\textbf{Analysis on the Learned Representations.} 
Further, we analyze the effect of DECL on representations. 
Similar to the pilot study, we induce some Gaussian noise to the raw data and compare the SNR value of the randomly initialized representations and that of the learned representations after optimization.
The result on the PTB-XL dataset is presented in Fig.~\ref{Fig_plot_wave}(d). 
We observe that the SNR value of a strong baseline CA-TCC almost remains steady after training, while the SNR value of our method is boosted significantly. 
See Appendix~\ref{appendx_exp_learned_embeddings} for detailed illustrations.
The analyses verify that DECL can indeed mitigate noise in representation learning.

\section{Related Work}

\subsection{Time Series Self-supervised Learning}

To alleviate the reliance on numerous annotated training data for supervised learning, an increasing number of researches focus on learning representations of time series in a self-supervised manner.
Time series SSL methods roughly fall into three categories~\cite{zhang2023self_SSL}: (1) contrastive-based methods~\cite{meng2023unsupervised}\cite{ma2023survey}\cite{MengQLCX023}, which are characterized by constructing positive and negative samples for contrastive learning, 
(2) generative-based methods~\cite{zerveas2021transformer}\cite{li2022generative}, which minimize the reconstruction error between raw data and the generated counterparts for model learning, 
(3) adversarial-based methods~\cite{seyfi2022generating}, which usually leverage generator and discriminator for adversarial learning. 
Among them, the contrastive-based methods are the dominant ones.
The key steps of contrastive learning involve building similar (positive) and dissimilar (negative) pairs of data samples and mapping the representations of positive pairs nearby while mapping those of negative pairs farther apart. 
The contrastive-based methods can be further divided into sampling-based~\cite{yeche2021neighborhood}\cite{tonekaboni2021unsupervised}, augmentation-based~\cite{yue2022ts2vec}\cite{yang2022unsupervised_icml}, and prediction-based~\cite{oord2018representation}\cite{deldari2021time}.
A recent method called TS-CoT~\cite{zhang2023co} assumes that complementary information from different views can be used to mitigate data noise.
Different from it, our method leverages the power of conventional denoising methods to guide representation learning.
More details are shown in Appendix~\ref{appendx_exp_realted_works}.

\subsection{Time Series Denoising}
Existing time series denoising methods can be divided into two categories: conventional methods and learning-based methods.
The conventional methods include (1) empirical
mode decomposition~\cite{huang1998empirical}, which decomposes the noisy data into a series of components and discards some of them to remove noise;
(2) wavelet filtering~\cite{chaovalit2011discrete}, which exploits wavelet transform, e.g., discrete wavelet transform, to remove the low amplitude coefficients associated with noise;
(3) sparse decomposition-based;
(4) Bayesian filtering~\cite{barber2011bayesian}, which relies on estimation theory to infer the noisy components, e.g., Kalman filter; 
(5) hybrid method, which involves a combination of these single methods (e.g., wavelet-based ICA method~\cite{castellanos2006recovering}) for noise mitigation.
The advantages of conventional methods lie in the great denoising effects if properly adopted to match the noise type. 
However, selecting suitable denoising methods requires prior knowledge or trial-and-error. 
To avoid human efforts, many learning-based methods that can mitigate the effect of noise have been proposed in recent years. 
Based on network architecture, these methods include wavelet neural networks, RNN-based~\cite{zhang2023robust}\cite{yoon2022robust}, and auto-encoders~\cite{zheng2022denoising}.
For example, D3VA~\cite{li2022generative} is a bidirectional variational auto-encoder that couples with diffusion and denoise modules for time series forecasting.
RLSTM~\cite{zhang2023robust} equips RNN models with localized stochastic sensitivity to alleviate the effect of noise for time series forecasting.
Differing from the previous efforts, our work leverages the conventional denoising methods to guide mitigating noise in learning, which combines the advantages of the two method categories.

\section{Conclusion}
In this work, we investigate the problem of mitigating the effect of data noise for time series SSL.
Accordingly, we propose an end-to-end method called
DEnoising-aware Contrastive Learning (DECL) for noise elimination in representation learning. 
It automatically selects suitable denoising methods for every sample to guide learning and performs a customized contrastive learning toward obtaining noise-free representations. 
Extensive empirical results verify the effectiveness of our method. 
Additionally, we perform comprehensive analyses to verify our claims, such as the distribution visualization of reconstruction errors and the denoising effect visualization of the selected methods.
Future works can explore (\textit{i}) how to automatically determine suitable hyper-parameters of the denoising methods; (\textit{ii}) examine the effectiveness of the method for more downstream tasks, e.g., forecasting and anomaly detection~\cite{lai2021revisiting}.

\section*{Ethical Statement}
There are no ethical issues.

\section*{Acknowledgments}
This research was supported by the grant of DaSAIL project P0030970 funded by PolyU (UGC).

\bibliographystyle{named}

\bibliography{ijcai24}

\begin{thebibliography}{}

\bibitem[\protect\citeauthoryear{Alday \bgroup \em et al.\egroup }{2020}]{alday2020classification}
Erick A~Perez Alday, Annie Gu, Amit~J Shah, Chad Robichaux, An-Kwok~Ian Wong, Chengyu Liu, Feifei Liu, Ali~Bahrami Rad, Andoni Elola, Salman Seyedi, et~al.
\newblock Classification of 12-lead ecgs: the physionet/computing in cardiology challenge 2020.
\newblock {\em Physiological measurement}, 41(12):124003, 2020.

\bibitem[\protect\citeauthoryear{Alickovic and Subasi}{2015}]{alickovic2015effect}
Emina Alickovic and Abdulhamit Subasi.
\newblock Effect of multiscale pca de-noising in ecg beat classification for diagnosis of cardiovascular diseases.
\newblock {\em Circuits, Systems, and Signal Processing}, 34:513--533, 2015.

\bibitem[\protect\citeauthoryear{Barber \bgroup \em et al.\egroup }{2011}]{barber2011bayesian}
David Barber, A~Taylan Cemgil, and Silvia Chiappa.
\newblock {\em Bayesian time series models}.
\newblock Cambridge University Press, 2011.

\bibitem[\protect\citeauthoryear{Burguera}{2018}]{burguera2018fast}
Antoni Burguera.
\newblock Fast qrs detection and ecg compression based on signal structural analysis.
\newblock {\em IEEE journal of biomedical and health informatics}, 23(1):123--131, 2018.

\bibitem[\protect\citeauthoryear{Castellanos and Makarov}{2006}]{castellanos2006recovering}
Nazareth~P Castellanos and Valeri~A Makarov.
\newblock Recovering eeg brain signals: Artifact suppression with wavelet enhanced independent component analysis.
\newblock {\em Journal of neuroscience methods}, 158(2):300--312, 2006.

\bibitem[\protect\citeauthoryear{Chandrakar \bgroup \em et al.\egroup }{2013}]{chandrakar2013survey}
Bhumika Chandrakar, OP~Yadav, and VK~Chandra.
\newblock A survey of noise removal techniques for ecg signals.
\newblock {\em International Journal of Advanced Research in Computer and Communication Engineering}, 2(3):1354--1357, 2013.

\bibitem[\protect\citeauthoryear{Chaovalit \bgroup \em et al.\egroup }{2011}]{chaovalit2011discrete}
Pimwadee Chaovalit, Aryya Gangopadhyay, George Karabatis, and Zhiyuan Chen.
\newblock Discrete wavelet transform-based time series analysis and mining.
\newblock {\em ACM Computing Surveys (CSUR)}, 43(2):1--37, 2011.

\bibitem[\protect\citeauthoryear{Chawla}{2011}]{chawla2011pca}
MPS Chawla.
\newblock Pca and ica processing methods for removal of artifacts and noise in electrocardiograms: A survey and comparison.
\newblock {\em Applied Soft Computing}, 11(2):2216--2226, 2011.

\bibitem[\protect\citeauthoryear{Chen \bgroup \em et al.\egroup }{2017}]{chen2017use}
Xun Chen, Xueyuan Xu, Aiping Liu, Martin~J McKeown, and Z~Jane Wang.
\newblock The use of multivariate emd and cca for denoising muscle artifacts from few-channel eeg recordings.
\newblock {\em IEEE transactions on instrumentation and measurement}, 67(2):359--370, 2017.

\bibitem[\protect\citeauthoryear{Chen \bgroup \em et al.\egroup }{2020}]{chen2020simple}
Ting Chen, Simon Kornblith, Mohammad Norouzi, and Geoffrey Hinton.
\newblock A simple framework for contrastive learning of visual representations.
\newblock In {\em ICML}, pages 1597--1607. PMLR, 2020.

\bibitem[\protect\citeauthoryear{Chowdhury \bgroup \em et al.\egroup }{2022}]{chowdhury2022tarnet}
Ranak~Roy Chowdhury, Xiyuan Zhang, Jingbo Shang, Rajesh~K Gupta, and Dezhi Hong.
\newblock Tarnet: Task-aware reconstruction for time-series transformer.
\newblock In {\em KDD}, pages 212--220, 2022.

\bibitem[\protect\citeauthoryear{Dau \bgroup \em et al.\egroup }{2019}]{dau2019ucr}
Hoang~Anh Dau, Anthony Bagnall, Kaveh Kamgar, Chin-Chia~Michael Yeh, Yan Zhu, Shaghayegh Gharghabi, Chotirat~Ann Ratanamahatana, and Eamonn Keogh.
\newblock The ucr time series archive.
\newblock {\em IEEE/CAA Journal of Automatica Sinica}, 6(6):1293--1305, 2019.

\bibitem[\protect\citeauthoryear{Deldari \bgroup \em et al.\egroup }{2021}]{deldari2021time}
Shohreh Deldari, Daniel~V Smith, Hao Xue, and Flora~D Salim.
\newblock Time series change point detection with self-supervised contrastive predictive coding.
\newblock In {\em Proceedings of the Web Conference}, pages 3124--3135, 2021.

\bibitem[\protect\citeauthoryear{Dera \bgroup \em et al.\egroup }{2023}]{dera2023trustworthy}
Dimah Dera, Sabeen Ahmed, Nidhal~C Bouaynaya, and Ghulam Rasool.
\newblock Trustworthy uncertainty propagation for sequential time-series analysis in rnns.
\newblock {\em IEEE Transactions on Knowledge and Data Engineering}, 2023.

\bibitem[\protect\citeauthoryear{Dong \bgroup \em et al.\egroup }{2023}]{dong2023simmtm}
Jiaxiang Dong, Haixu Wu, Haoran Zhang, Li~Zhang, Jianmin Wang, and Mingsheng Long.
\newblock Simmtm: A simple pre-training framework for masked time-series modeling.
\newblock In {\em NeurIPS}, 2023.

\bibitem[\protect\citeauthoryear{Eldele \bgroup \em et al.\egroup }{2021}]{eldele2021time}
Emadeldeen Eldele, Mohamed Ragab, Zhenghua Chen, Min Wu, Chee~Keong Kwoh, Xiaoli Li, and Cuntai Guan.
\newblock Time-series representation learning via temporal and contextual contrasting.
\newblock {\em IJCAI}, 2021.

\bibitem[\protect\citeauthoryear{Eldele \bgroup \em et al.\egroup }{2023}]{eldele2023self}
Emadeldeen Eldele, Mohamed Ragab, Zhenghua Chen, Min Wu, Chee-Keong Kwoh, Xiaoli Li, and Cuntai Guan.
\newblock Self-supervised contrastive representation learning for semi-supervised time-series classification.
\newblock {\em IEEE Transactions on Pattern Analysis and Machine Intelligence}, 2023.

\bibitem[\protect\citeauthoryear{Fang \bgroup \em et al.\egroup }{2021}]{fang2021dual}
Bo~Fang, Junxin Chen, Yu~Liu, Wei Wang, Ke~Wang, Amit~Kumar Singh, and Zhihan Lv.
\newblock Dual-channel neural network for atrial fibrillation detection from a single lead ecg wave.
\newblock {\em IEEE journal of biomedical and health informatics}, 2021.

\bibitem[\protect\citeauthoryear{Goldberger \bgroup \em et al.\egroup }{2000}]{goldberger2000physiobank}
Ary~L Goldberger, Luis~AN Amaral, Leon Glass, Jeffrey~M Hausdorff, Plamen~Ch Ivanov, Roger~G Mark, Joseph~E Mietus, George~B Moody, Chung-Kang Peng, and H~Eugene Stanley.
\newblock Physiobank, physiotoolkit, and physionet: components of a new research resource for complex physiologic signals.
\newblock {\em Circulation}, 101(23):e215--e220, 2000.

\bibitem[\protect\citeauthoryear{He \bgroup \em et al.\egroup }{2015}]{he2015optimal}
Hong He, Yonghong Tan, and Yuexia Wang.
\newblock Optimal base wavelet selection for ecg noise reduction using a comprehensive entropy criterion.
\newblock {\em Entropy}, 17(9):6093--6109, 2015.

\bibitem[\protect\citeauthoryear{Huang \bgroup \em et al.\egroup }{1998}]{huang1998empirical}
Norden~E Huang, Zheng Shen, Steven~R Long, Manli~C Wu, Hsing~H Shih, Quanan Zheng, Nai-Chyuan Yen, Chi~Chao Tung, and Henry~H Liu.
\newblock The empirical mode decomposition and the hilbert spectrum for nonlinear and non-stationary time series analysis.
\newblock {\em Proceedings of the Royal Society of London. Series A: mathematical, physical and engineering sciences}, 454(1971):903--995, 1998.

\bibitem[\protect\citeauthoryear{Ismail~Fawaz \bgroup \em et al.\egroup }{2019}]{ismail2019deep}
Hassan Ismail~Fawaz, Germain Forestier, Jonathan Weber, Lhassane Idoumghar, and Pierre-Alain Muller.
\newblock Deep learning for time series classification: a review.
\newblock {\em Data mining and knowledge discovery}, 33(4):917--963, 2019.

\bibitem[\protect\citeauthoryear{Jeha \bgroup \em et al.\egroup }{2021}]{jeha2021psa}
Paul Jeha, Michael Bohlke-Schneider, Pedro Mercado, Shubham Kapoor, Rajbir~Singh Nirwan, Valentin Flunkert, Jan Gasthaus, and Tim Januschowski.
\newblock Psa-gan: Progressive self attention gans for synthetic time series.
\newblock In {\em ICLR}, 2021.

\bibitem[\protect\citeauthoryear{Jeon \bgroup \em et al.\egroup }{2022}]{jeon2022gt}
Jinsung Jeon, Jeonghak Kim, Haryong Song, Seunghyeon Cho, and Noseong Park.
\newblock Gt-gan: General purpose time series synthesis with generative adversarial networks.
\newblock {\em NeurIPS}, 35:36999--37010, 2022.

\bibitem[\protect\citeauthoryear{Kiyasseh \bgroup \em et al.\egroup }{2021}]{kiyasseh2021clocs}
Dani Kiyasseh, Tingting Zhu, and David~A Clifton.
\newblock Clocs: Contrastive learning of cardiac signals across space, time, and patients.
\newblock In {\em ICML}, 2021.

\bibitem[\protect\citeauthoryear{Kotte and Dabbakuti}{2020}]{kotte2020methods}
Shailaja Kotte and JRK~Kumar Dabbakuti.
\newblock Methods for removal of artifacts from eeg signal: A review.
\newblock In {\em Journal of Physics: Conference Series}, volume 1706, page 012093. IOP Publishing, 2020.

\bibitem[\protect\citeauthoryear{Lai \bgroup \em et al.\egroup }{2021}]{lai2021revisiting}
Kwei-Herng Lai, Daochen Zha, Junjie Xu, Yue Zhao, Guanchu Wang, and Xia Hu.
\newblock Revisiting time series outlier detection: Definitions and benchmarks.
\newblock In {\em NeurIPS}, 2021.

\bibitem[\protect\citeauthoryear{Lai \bgroup \em et al.\egroup }{2023}]{lai2023practical}
Jiewei Lai, Huixin Tan, Jinliang Wang, Lei Ji, Jun Guo, Baoshi Han, Yajun Shi, Qianjin Feng, and Wei Yang.
\newblock Practical intelligent diagnostic algorithm for wearable 12-lead ecg via self-supervised learning on large-scale dataset.
\newblock {\em Nature Communications}, 14(1):3741, 2023.

\bibitem[\protect\citeauthoryear{Lan \bgroup \em et al.\egroup }{2022}]{lan2022intra}
Xiang Lan, Dianwen Ng, Shenda Hong, and Mengling Feng.
\newblock Intra-inter subject self-supervised learning for multivariate cardiac signals.
\newblock In {\em AAAI}, 2022.

\bibitem[\protect\citeauthoryear{Lessmeier \bgroup \em et al.\egroup }{2016}]{lessmeier2016condition}
Christian Lessmeier, James~Kuria Kimotho, Detmar Zimmer, and Walter Sextro.
\newblock Condition monitoring of bearing damage in electromechanical drive systems by using motor current signals of electric motors: A benchmark data set for data-driven classification.
\newblock In {\em PHM Society European Conference}, volume~3, 2016.

\bibitem[\protect\citeauthoryear{Li and Marlin}{2020}]{li2020learning}
Steven Cheng-Xian Li and Benjamin Marlin.
\newblock Learning from irregularly-sampled time series: A missing data perspective.
\newblock In {\em ICML}, pages 5937--5946. PMLR, 2020.

\bibitem[\protect\citeauthoryear{Li \bgroup \em et al.\egroup }{2022}]{li2022generative}
Yan Li, Xinjiang Lu, Yaqing Wang, and Dejing Dou.
\newblock Generative time series forecasting with diffusion, denoise, and disentanglement.
\newblock {\em NeurIPS}, 2022.

\bibitem[\protect\citeauthoryear{Liu \bgroup \em et al.\egroup }{2018}]{liu2018open}
Feifei Liu, Chengyu Liu, Lina Zhao, Xiangyu Zhang, Xiaoling Wu, Xiaoyan Xu, Yulin Liu, Caiyun Ma, Shoushui Wei, Zhiqiang He, et~al.
\newblock An open access database for evaluating the algorithms of electrocardiogram rhythm and morphology abnormality detection.
\newblock {\em Journal of Medical Imaging and Health Informatics}, 8(7):1368--1373, 2018.

\bibitem[\protect\citeauthoryear{Lu \bgroup \em et al.\egroup }{2009}]{lu2009model}
Yan Lu, Jingyu Yan, and Yeung Yam.
\newblock Model-based ecg denoising using empirical mode decomposition.
\newblock In {\em BIBM}, pages 191--196. IEEE, 2009.

\bibitem[\protect\citeauthoryear{Luo \bgroup \em et al.\egroup }{2019}]{luo2019e2gan}
Yonghong Luo, Ying Zhang, Xiangrui Cai, and Xiaojie Yuan.
\newblock E2gan: End-to-end generative adversarial network for multivariate time series imputation.
\newblock In {\em AAAI}, 2019.

\bibitem[\protect\citeauthoryear{Ma \bgroup \em et al.\egroup }{2023}]{ma2023survey}
Qianli Ma, Zhen Liu, Zhenjing Zheng, Ziyang Huang, Siying Zhu, Zhongzhong Yu, and James~T Kwok.
\newblock A survey on time-series pre-trained models.
\newblock {\em arXiv}, 2023.

\bibitem[\protect\citeauthoryear{Meng \bgroup \em et al.\egroup }{2023a}]{MengQLCX023}
Qianwen Meng, Hangwei Qian, Yong Liu, Lizhen Cui, Yonghui Xu, and Zhiqi Shen.
\newblock {MHCCL:} masked hierarchical cluster-wise contrastive learning for multivariate time series.
\newblock In {\em AAAI}, pages 9153--9161, 2023.

\bibitem[\protect\citeauthoryear{Meng \bgroup \em et al.\egroup }{2023b}]{meng2023unsupervised}
Qianwen Meng, Hangwei Qian, Yong Liu, Yonghui Xu, Zhiqi Shen, and Lizhen Cui.
\newblock Unsupervised representation learning for time series: A review.
\newblock {\em arXiv}, 2023.

\bibitem[\protect\citeauthoryear{Mneimneh \bgroup \em et al.\egroup }{2006}]{mneimneh2006adaptive}
MA~Mneimneh, EE~Yaz, MT~Johnson, and RJ~Povinelli.
\newblock An adaptive kalman filter for removing baseline wandering in ecg signals.
\newblock In {\em 2006 Computers in Cardiology}, pages 253--256. IEEE, 2006.

\bibitem[\protect\citeauthoryear{Nonnenmacher \bgroup \em et al.\egroup }{2022}]{nonnenmacher2022utilizing}
Manuel~T Nonnenmacher, Lukas Oldenburg, Ingo Steinwart, and David Reeb.
\newblock Utilizing expert features for contrastive learning of time-series representations.
\newblock In {\em ICML}, pages 16969--16989. PMLR, 2022.

\bibitem[\protect\citeauthoryear{Oord \bgroup \em et al.\egroup }{2018}]{oord2018representation}
Aaron van~den Oord, Yazhe Li, and Oriol Vinyals.
\newblock Representation learning with contrastive predictive coding.
\newblock {\em arXiv}, 2018.

\bibitem[\protect\citeauthoryear{Prabhakararao and Dandapat}{2022}]{PrabhakararaoD22}
Eedara Prabhakararao and Samarendra Dandapat.
\newblock Multi-scale convolutional neural network ensemble for multi-class arrhythmia classification.
\newblock {\em IEEE Journal of Biomedical and Health Informatics}, 26(8):3802--3812, 2022.

\bibitem[\protect\citeauthoryear{Robbins \bgroup \em et al.\egroup }{2020}]{robbins2020sensitive}
Kay~A Robbins, Jonathan Touryan, Tim Mullen, Christian Kothe, and Nima Bigdely-Shamlo.
\newblock How sensitive are eeg results to preprocessing methods: a benchmarking study.
\newblock {\em IEEE transactions on neural systems and rehabilitation engineering}, 28(5):1081--1090, 2020.

\bibitem[\protect\citeauthoryear{Romero}{2011}]{romero2011pca}
I~Romero.
\newblock Pca and ica applied to noise reduction in multi-lead ecg.
\newblock In {\em 2011 Computing in Cardiology}, pages 613--616. IEEE, 2011.

\bibitem[\protect\citeauthoryear{Schneider \bgroup \em et al.\egroup }{2022}]{schneider2022detecting}
Tim Schneider, Chen Qiu, Marius Kloft, Decky~Aspandi Latif, Steffen Staab, Stephan Mandt, and Maja Rudolph.
\newblock Detecting anomalies within time series using local neural transformations.
\newblock {\em arXiv}, 2022.

\bibitem[\protect\citeauthoryear{Seyfi \bgroup \em et al.\egroup }{2022}]{seyfi2022generating}
Ali Seyfi, Jean-Francois Rajotte, and Raymond Ng.
\newblock Generating multivariate time series with common source coordinated gan (cosci-gan).
\newblock {\em NeurIPS}, 35:32777--32788, 2022.

\bibitem[\protect\citeauthoryear{Singh \bgroup \em et al.\egroup }{2016}]{singh2016comparative}
Vivek Singh, Karan Veer, Reecha Sharma, and Sanjeev Kumar.
\newblock Comparative study of fir and iir filters for the removal of 50 hz noise from eeg signal.
\newblock {\em International Journal of Biomedical Engineering and Technology}, 22(3):250--257, 2016.

\bibitem[\protect\citeauthoryear{Tonekaboni \bgroup \em et al.\egroup }{2021}]{tonekaboni2021unsupervised}
Sana Tonekaboni, Danny Eytan, and Anna Goldenberg.
\newblock Unsupervised representation learning for time series with temporal neighborhood coding.
\newblock {\em ICLR}, 2021.

\bibitem[\protect\citeauthoryear{Vaid \bgroup \em et al.\egroup }{2023}]{vaid2023foundational}
Akhil Vaid, Joy Jiang, Ashwin Sawant, Stamatios Lerakis, Edgar Argulian, Yuri Ahuja, Joshua Lampert, Alexander Charney, Hayit Greenspan, Jagat Narula, et~al.
\newblock A foundational vision transformer improves diagnostic performance for electrocardiograms.
\newblock {\em NPJ Digital Medicine}, 6(1):108, 2023.

\bibitem[\protect\citeauthoryear{Vaswani \bgroup \em et al.\egroup }{2017}]{vaswani2017attention}
Ashish Vaswani, Noam Shazeer, Niki Parmar, Jakob Uszkoreit, Llion Jones, Aidan~N Gomez, {\L}ukasz Kaiser, and Illia Polosukhin.
\newblock Attention is all you need.
\newblock {\em NeurIPS}, 30, 2017.

\bibitem[\protect\citeauthoryear{Vu \bgroup \em et al.\egroup }{2017}]{Vu2017DataEI}
Van~Tam Vu, Duc-Tan Tran, and Trong~Hanh Phan.
\newblock Data embedding in audio signal using multiple bit marking layers method.
\newblock {\em Multimedia Tools and Applications}, 76:11391--11406, 2017.

\bibitem[\protect\citeauthoryear{Wagner \bgroup \em et al.\egroup }{2020}]{wagner2020ptb}
Patrick Wagner, Nils Strodthoff, Ralf-Dieter Bousseljot, Dieter Kreiseler, Fatima~I Lunze, Wojciech Samek, and Tobias Schaeffter.
\newblock Ptb-xl, a large publicly available electrocardiography dataset.
\newblock {\em Scientific data}, 7(1):154, 2020.

\bibitem[\protect\citeauthoryear{Wang \bgroup \em et al.\egroup }{2020}]{Wang2020DeepMF}
Ruxin Wang, Jianping Fan, and Ye~Li.
\newblock Deep multi-scale fusion neural network for multi-class arrhythmia detection.
\newblock {\em IEEE Journal of Biomedical and Health Informatics}, 24:2461--2472, 2020.

\bibitem[\protect\citeauthoryear{Wang \bgroup \em et al.\egroup }{2022}]{wang2022learning}
Zhiyuan Wang, Xovee Xu, Weifeng Zhang, Goce Trajcevski, Ting Zhong, and Fan Zhou.
\newblock Learning latent seasonal-trend representations for time series forecasting.
\newblock {\em NeurIPS}, 35:38775--38787, 2022.

\bibitem[\protect\citeauthoryear{Wu \bgroup \em et al.\egroup }{2009}]{wu2009filtering}
Yunfeng Wu, Rangaraj~M Rangayyan, Yachao Zhou, and Sin-Chun Ng.
\newblock Filtering electrocardiographic signals using an unbiased and normalized adaptive noise reduction system.
\newblock {\em Medical Engineering \& Physics}, 31(1):17--26, 2009.

\bibitem[\protect\citeauthoryear{Yang and Hong}{2022}]{yang2022unsupervised_icml}
Ling Yang and Shenda Hong.
\newblock Unsupervised time-series representation learning with iterative bilinear temporal-spectral fusion.
\newblock In {\em ICML}, 2022.

\bibitem[\protect\citeauthoryear{Yang \bgroup \em et al.\egroup }{2022}]{yang2022timeclr}
Xinyu Yang, Zhenguo Zhang, and Rongyi Cui.
\newblock Timeclr: A self-supervised contrastive learning framework for univariate time series representation.
\newblock {\em Knowledge-Based Systems}, 245:108606, 2022.

\bibitem[\protect\citeauthoryear{Yang \bgroup \em et al.\egroup }{2023}]{yang2023biot}
Chaoqi Yang, M~Brandon Westover, and Jimeng Sun.
\newblock Biot: Biosignal transformer for cross-data learning in the wild.
\newblock In {\em NeurIPS}, 2023.

\bibitem[\protect\citeauthoryear{Y{\`e}che \bgroup \em et al.\egroup }{2021}]{yeche2021neighborhood}
Hugo Y{\`e}che, Gideon Dresdner, Francesco Locatello, Matthias H{\"u}ser, and Gunnar R{\"a}tsch.
\newblock Neighborhood contrastive learning applied to online patient monitoring.
\newblock In {\em ICML}, pages 11964--11974, 2021.

\bibitem[\protect\citeauthoryear{Yoon \bgroup \em et al.\egroup }{2022}]{yoon2022robust}
TaeHo Yoon, Youngsuk Park, Ernest~K Ryu, and Yuyang Wang.
\newblock Robust probabilistic time series forecasting.
\newblock In {\em AISTATS}, pages 1336--1358. PMLR, 2022.

\bibitem[\protect\citeauthoryear{Yue \bgroup \em et al.\egroup }{2022}]{yue2022ts2vec}
Zhihan Yue, Yujing Wang, Juanyong Duan, Tianmeng Yang, Congrui Huang, Yunhai Tong, and Bixiong Xu.
\newblock Ts2vec: Towards universal representation of time series.
\newblock In {\em AAAI}, volume~36, pages 8980--8987, 2022.

\bibitem[\protect\citeauthoryear{Zerveas \bgroup \em et al.\egroup }{2021}]{zerveas2021transformer}
George Zerveas, Srideepika Jayaraman, Dhaval Patel, Anuradha Bhamidipaty, and Carsten Eickhoff.
\newblock A transformer-based framework for multivariate time series representation learning.
\newblock In {\em KDD}, pages 2114--2124, 2021.

\bibitem[\protect\citeauthoryear{Zhang \bgroup \em et al.\egroup }{2021a}]{zhang2021eegdenoisenet}
Haoming Zhang, Mingqi Zhao, Chen Wei, Dante Mantini, Zherui Li, and Quanying Liu.
\newblock Eegdenoisenet: a benchmark dataset for deep learning solutions of eeg denoising.
\newblock {\em Journal of Neural Engineering}, 18(5):056057, 2021.

\bibitem[\protect\citeauthoryear{Zhang \bgroup \em et al.\egroup }{2021b}]{zhang2021heartbeats}
Yatao Zhang, Junyan Li, Shoushui Wei, Fengyu Zhou, and Dong Li.
\newblock Heartbeats classification using hybrid time-frequency analysis and transfer learning based on resnet.
\newblock {\em IEEE Journal of Biomedical and Health Informatics}, 25(11):4175--4184, 2021.

\bibitem[\protect\citeauthoryear{Zhang \bgroup \em et al.\egroup }{2022}]{zhang2022self}
Xiang Zhang, Ziyuan Zhao, Theodoros Tsiligkaridis, and Marinka Zitnik.
\newblock Self-supervised contrastive pre-training for time series via time-frequency consistency.
\newblock {\em NeurIPS}, 35:3988--4003, 2022.

\bibitem[\protect\citeauthoryear{Zhang \bgroup \em et al.\egroup }{2023a}]{zhang2023self_SSL}
Kexin Zhang, Qingsong Wen, Chaoli Zhang, Rongyao Cai, Ming Jin, Yong Liu, James Zhang, Yuxuan Liang, Guansong Pang, Dongjin Song, et~al.
\newblock Self-supervised learning for time series analysis: Taxonomy, progress, and prospects.
\newblock {\em arXiv}, 2023.

\bibitem[\protect\citeauthoryear{Zhang \bgroup \em et al.\egroup }{2023b}]{zhang2023co}
Weiqi Zhang, Jianfeng Zhang, Jia Li, and Fugee Tsung.
\newblock A co-training approach for noisy time series learning.
\newblock In {\em CIKM}, pages 3308--3318, 2023.

\bibitem[\protect\citeauthoryear{Zhang \bgroup \em et al.\egroup }{2023c}]{zhang2023self}
Wenrui Zhang, Ling Yang, Shijia Geng, and Shenda Hong.
\newblock Self-supervised time series representation learning via cross reconstruction transformer.
\newblock {\em IEEE Transactions on Neural Networks and Learning Systems}, 2023.

\bibitem[\protect\citeauthoryear{Zhang \bgroup \em et al.\egroup }{2023d}]{zhang2023robust}
Xueli Zhang, Cankun Zhong, Jianjun Zhang, Ting Wang, and Wing~WY Ng.
\newblock Robust recurrent neural networks for time series forecasting.
\newblock {\em Neurocomputing}, 526:143--157, 2023.

\bibitem[\protect\citeauthoryear{Zheng \bgroup \em et al.\egroup }{2020}]{zheng202012}
Jianwei Zheng, Jianming Zhang, Sidy Danioko, Hai Yao, Hangyuan Guo, and Cyril Rakovski.
\newblock A 12-lead electrocardiogram database for arrhythmia research covering more than 10,000 patients.
\newblock {\em Scientific data}, 7(1):48, 2020.

\bibitem[\protect\citeauthoryear{Zheng \bgroup \em et al.\egroup }{2022}]{zheng2022denoising}
Zhong Zheng, Zijun Zhang, Long Wang, and Xiong Luo.
\newblock Denoising temporal convolutional recurrent autoencoders for time series classification.
\newblock {\em Information Sciences}, 588:159--173, 2022.

\end{thebibliography}

\appendix

\section{Appendix}
Details on algorithm design, experiment setting, and additional results.

\subsection{Datasets}\label{Datasets}
\begin{itemize}

\item \textbf{SleepEDF}~\cite{goldberger2000physiobank} contains $153$ whole-night sleeping electroencephalogram (EEG) records from $82$ subjects. We further process the data and obtain single-channel EEG signals with $100$ Hz. The records encompass five distinct sleep phases: wake (W), non-rapid eye movement (N1, N2, N3), and rapid eye movement (REM).

\item \textbf{FaultDiagnosis}~\cite{lessmeier2016condition} is compiled using sensor measurements from a bearing machine with four distinct working conditions. All the records fall into three classes: two fault classes (i.e., inner fault and outer fault) and one normal class.

\item \textbf{CPSC18}~\cite{liu2018open}
is collected from $9,458$ patients and the electrocardiogram (ECG) records duration ranges from $6$ to $60$ seconds. 
We resample the data at $200$ Hz, ensuring each record has a fixed duration of $10$ seconds. This involves cropping longer ECG records and padding shorter ones with zeros. Additionally, only single-label records are retained.
The curated dataset comprises nine diagnostic categories: normal sinus rhythm (Normal), atrial fibrillation (AF), premature atrial contraction (PAC), first-degree atrioventricular block (IAVB), right bundle branch block (RBBB), ST-segment depression (STD), premature ventricular contraction (PVC), left bundle branch block (LBBB), and ST-segment elevation (STE).

\item \textbf{PTB-XL}~\cite{wagner2020ptb}
is a publicly accessible ECG dataset from $18,885$ patients. 
We resample the data with $200$ Hz, standardizing all records to a duration of $10$ seconds. 
The dataset includes a normal ECG category and four arrhythmia categories: myocardial infarction (MI), hypertrophy (HYP), ST/T changes (STTC), and conduction disturbance (CD).

\item \textbf{Georgia}~\cite{alday2020classification} This ECG dataset represents a distinct demographic from the Southeastern United States. 
We resample the recordings at $200$ Hz and standardize each to a $10$-second duration. 
Only ECG types with over $100$ records are retained, including normal rhythm, sinus bradycardia (SB), sinus tachycardia (STach), nonspecific ST-T abnormality (NSTTA), T wave abnormality (TAB), left ventricular hypertrophy (LVH), and sinus arrhythmia (SA).

\end{itemize}
Furthermore, given that the processed CPSC18 and Georgia datasets have limited data in specific classes, we perform data augmentation by following related studies~\cite{PrabhakararaoD22}\cite{Wang2020DeepMF}. Specifically, we scale up the ECG records for each class by varying the scaling degree \{$1.02$, $1.04$, $1.06$, $1.08$\}, thus expanding the datasets four times.

\subsection{Comparative SSL Methods}\label{SSL_Methods}

We compare our method with the following baselines.

\begin{itemize}

\item \textbf{TF-C}~\cite{zhang2022self} is a self-supervised learning (SSL) method that leverages the consistency of time series between time and frequency domains for contrastive learning.

\item \textbf{TS2vec}~\cite{yue2022ts2vec} is a contrastive learning-based method that hierarchically discriminates positive and negative samples at instance-wise and temporal dimensions.

\item \textbf{CRT}~\cite{zhang2023self} is a novel method that captures the temporal-spectral correlations of time series for representation learning. It first converts data into the frequency domain, then drops certain patches in both the time and frequency domains, and exploits a transformer for reconstruction learning. 

\item \textbf{SimMTM}~\cite{dong2023simmtm} is an unsupervised method with masked reconstruction for representation learning. Specifically, it masks a random portion of the input data and recovers them by the weighted aggregation of multiple neighbors outside the manifold.

\item \textbf{TS-CoT}~\cite{zhang2023co} combines co-training and contrastive learning to mitigate the impact of data noise for representation learning. 
Specifically, it builds two views for the input data via two different encoders and exploits the complementary information for learning.

\item \textbf{CA-TCC}~\cite{eldele2023self} is an advanced SSL method for time series data. It first transforms raw data into two views via weak and strong augmentations and then maximizes the similarity among different views of the sample while minimizing the similarity among views of different samples for contrastive learning.

\end{itemize}

\section{Additional Experimental Analysis}

\subsection{Additional Results on Linear Evaluation of Representations}\label{appendx_exp_linear_eval}
As discussed in §$4.3$, we use linear evaluation to examine the effectiveness of the learned representations. Specifically, we pre-train SSL methods with unlabeled data for representation learning, then use a portion of labeled data for learning a linear classifier. 
The results with $30$\% training labels are provided in Table~\ref{Table_1_overall_appedx}.

\begin{table*}[t]
\centering
\caption{Overall performance (\%) with $30$ percent of training labels in the linear evaluation. 
}
\resizebox{0.95\linewidth}{!}{
\begin{tabular}{ccccccccccc}
\toprule
\multirow{2}{*}{\textbf{Methods}} & \multicolumn{2}{c}{\textbf{SleepEDF}} & \multicolumn{2}{c}{\textbf{FaultDiagnosis}} & \multicolumn{2}{c}{\textbf{CPSC18}} & \multicolumn{2}{c}{\textbf{PTB-XL}} & \multicolumn{2}{c}{\textbf{Georgia}} \\
\cmidrule(r){2-3} \cmidrule(r){4-5} \cmidrule(r){6-7} \cmidrule(r){8-9} \cmidrule(r){10-11}
                                  & \textit{Accuracy}         & $\textit{Weighted-F}_{1}$        & \textit{Accuracy}         & $\textit{Weighted-F}_{1}$        & \textit{Accuracy}        & $\textit{Weighted-F}_{1}$       & \textit{Accuracy}        & $\textit{Weighted-F}_{1}$       & \textit{Accuracy}        & $\textit{Weighted-F}_{1}$        \\
\midrule
TF-C                              & 68.23±0.67        & 67.85±0.56           & 79.03±0.76          & 78.41±0.61            & 41.34±0.62        & 37.96±0.73           & 52.31±0.69        & 43.85±0.43           & 51.82±0.46        & 36.73±0.49           \\
TF-C + \textit{DN}                         & 71.45±0.73        & 71.17±0.64           & 81.94±0.69          & 81.63±0.82            & 43.05±0.44        & 40.24±0.58           & 54.87±0.52        & 46.38±0.65           & 53.79±0.53        & 38.95±0.42           \\
TF-C + \textit{Merge}                       & 68.92±0.62        & 68.26±0.51           & 79.65±0.95          & 79.13±0.70            & 41.68±0.63        & 38.70±0.57           & 53.62±0.58        & 44.76±0.41           & 52.26±0.68        & 37.39±0.57           \\
TS2vec                            & 68.91±0.68        & 68.47±0.37           & 80.58±0.62          & 80.22±0.49            & 42.36±0.65        & 39.72±0.47           & 53.52±0.46        & 45.12±0.32           & 53.18±0.52        & 39.34±0.36           \\
TS2vec + \textit{DN}                       & 73.49±0.51        & 73.14±0.43           & 84.86±0.43          & 84.59±0.34            & 44.61±0.59        & 41.85±0.43           & 55.94±0.35        & 47.23±0.27           & 55.07±0.29        & 41.72±0.23           \\
TS2vec + \textit{Merge}                     & 70.51±0.66        & 69.95±0.52           & 82.39±0.71          & 81.84±0.65            & 42.96±0.86        & 40.23±0.71           & 53.80±0.47        & 45.46±0.36           & 54.24±0.45        & 40.31±0.38           \\
CRT                               & 67.93±0.54        & 67.16±0.40           & 78.92±0.74          & 78.50±0.46            & 41.05±0.74        & 38.85±0.45           & 53.48±0.54        & 44.09±0.38           & 51.83±0.34        & 37.02±0.23           \\
CRT + \textit{DN}                          & 71.65±0.47        & 71.23±0.35           & 82.17±0.65          & 82.36±0.43            & 43.38±0.56        & 40.62±0.39           & 55.69±0.39        & 46.34±0.19           & 54.14±0.31        & 39.48±0.25           \\
CRT + \textit{Merge}                        & 68.42±0.58        & 68.32±0.41           & 79.85±0.68          & 79.22±0.51            & 41.71±0.81        & 39.58±0.76           & 53.94±0.43        & 45.96±0.26           & 52.75±0.37        & 37.89±0.28           \\
SimMTM                            & 69.89±0.63        & 68.72±0.53           & 82.10±0.57          & 81.74±0.44            & 42.96±0.62        & 39.87±0.43           & 53.75±0.45        & 45.81±0.31           & 53.56±0.33        & 39.83±0.41           \\
SimMTM + \textit{DN}                       & 74.73±0.44        & 74.05±0.49           & 85.46±0.42          & 85.21±0.19            & 45.03±0.41        & 42.29±0.22           & 56.13±0.40        & 47.95±0.12           & 55.32±0.36        & 42.29±0.43           \\
SimMTM + \textit{Merge}                     & 70.58±0.59        & 69.81±0.38           & 83.34±0.41          & 82.75±0.36            & 43.67±0.73        & 41.36±0.49           & 54.80±0.41        & 46.71±0.37           & 54.69±0.39        & 40.58±0.37           \\
TS-CoT                            & 68.49±0.65        & 68.12±0.34           & 80.61±0.54          & 79.72±0.42            & 41.28±0.79        & 39.45±0.54           & 53.04±0.51        & 44.13±0.24           & 52.53±0.41        & 37.84±0.45           \\
TS-CoT + \textit{DN}                       & 72.18±0.53        & 72.79±0.25           & 83.83±0.46          & 83.41±0.28            & 43.74±0.63        & 41.34±0.37           & 55.37±0.36        & 46.70±0.23           & 54.64±0.32        & 40.17±0.14           \\
TS-CoT + \textit{Merge}                     & 69.35±0.57        & 68.74±0.47           & 81.33±0.67          & 80.96±0.43            & 42.89±0.92        & 40.62±0.51           & 54.74±0.62        & 44.92±0.35           & 53.42±0.58        & 38.63±0.36           \\
CA-TCC                            & 70.56±0.49        & 69.73±0.36           & 82.15±0.58          & 81.73±0.37            & 42.64±0.57        & 40.81±0.46           & 53.98±0.37        & 46.27±0.26           & 53.49±0.30        & 40.16±0.39           \\
CA-TCC + \textit{DN}                       & 74.69±0.32        & 73.94±0.29           & 85.69±0.31          & 85.36±0.24            & 45.39±0.48        & 42.47±0.29           & 56.21±0.29        & 47.62±0.17           & 55.18±0.26        & 42.46±0.24           \\
CA-TCC + \textit{Merge}                     & 71.18±0.54        & 70.74±0.43           & 83.41±0.47          & 82.84±0.39            & 43.46±0.73        & 41.12±0.64           & 54.46±0.46        & 46.53±0.38           & 53.61±0.35        & 40.75±0.38           \\
\textbf{DECL (Ours)}               & \textbf{77.21±0.47}        & \textbf{76.75±0.35}           & \textbf{87.94±0.39}          & \textbf{87.52±0.28}            & \textbf{48.03±0.36}        & \textbf{44.90±0.43}           & \textbf{59.73±0.45}        & \textbf{51.14±0.32}           & \textbf{57.82±0.37}        & \textbf{44.71±0.32}           \\
\bottomrule
\end{tabular}
}
\label{Table_1_overall_appedx}
\end{table*}

\subsection{Robustness Analysis against Data Noise}\label{appendx_exp_robustness}
\begin{figure}[t]
\begin{center}
\begin{tabular}{c}
\includegraphics[width = 0.95\linewidth]{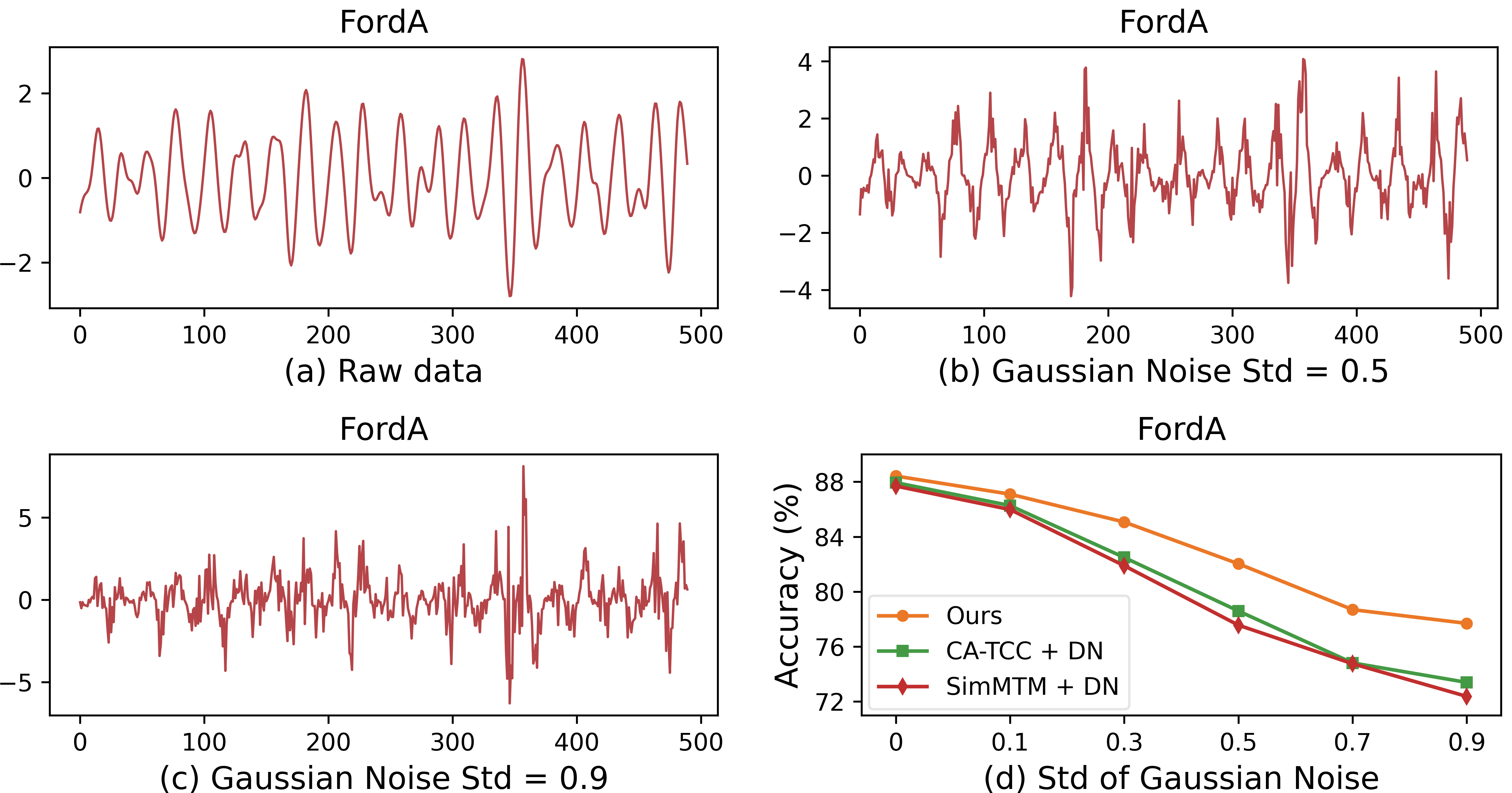}
\end{tabular}
\end{center}
\caption{Robustness analysis results on FordA dataset.}
\label{Fig_robustness_FordA_Appdix}
\end{figure}

In addition to the results on the noisy datasets in §$4.6$, we perform robustness analysis on the FordA dataset~\cite{dau2019ucr} that is of trivial data noise. The results are shown in Fig.~\ref{Fig_robustness_FordA_Appdix}.
We observe that our method achieves comparable performance with state-of-the-art SSL methods (e.g., CA-TCC and SimMTM) when the data suffer from trivial noise (shown in Fig.~\ref{Fig_robustness_FordA_Appdix}(a)). 
As the data noises rise, the performance of the methods gradually declines, and more importantly, our method consistently outperforms the strong SSL methods.

\begin{figure}[t]
\begin{center}
\begin{tabular}{c}
\includegraphics[width = 0.95\linewidth]{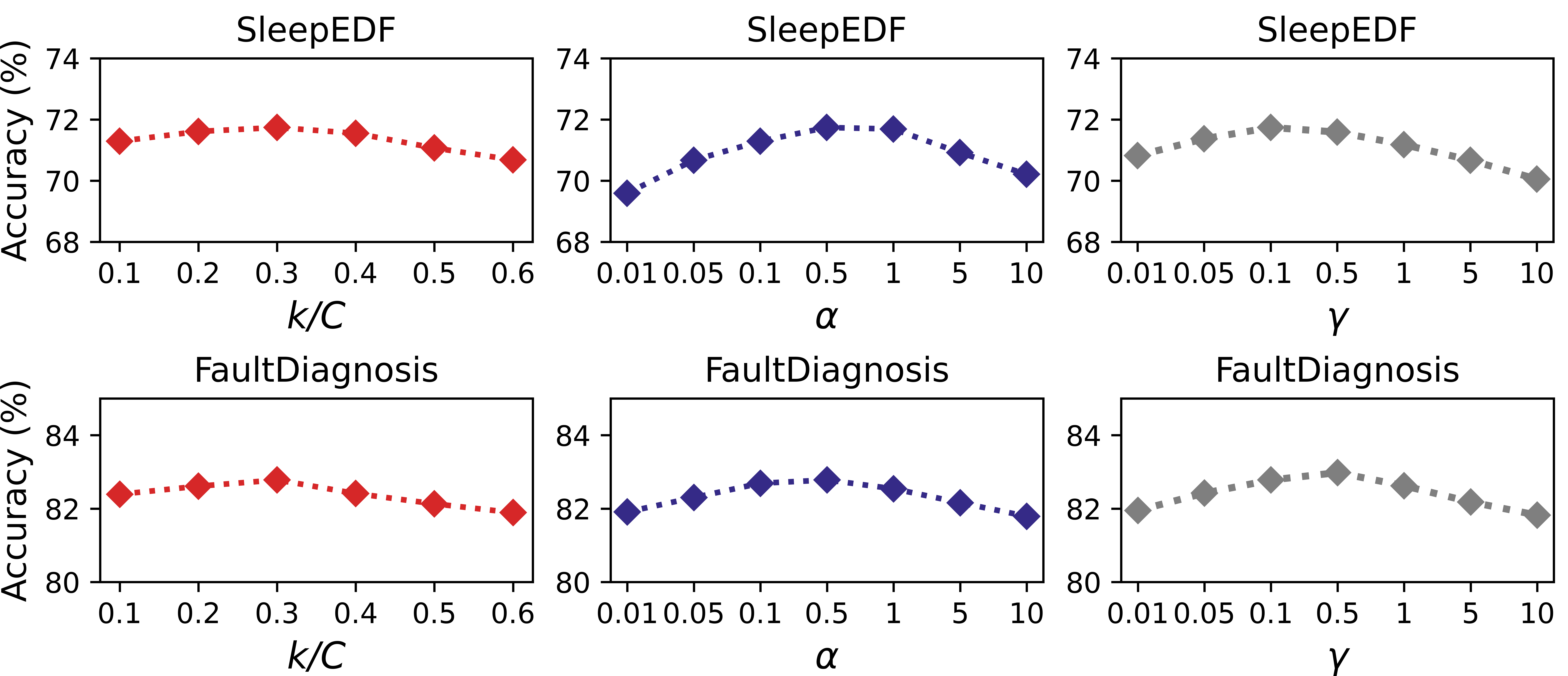}
\end{tabular}
\end{center}
\caption{Hyper-parameter analysis on other datasets.}
\label{Fig_hyperparam_Appdix}
\end{figure}

\subsection{Hyper-parameter Analysis}\label{appendx_exp_hyperparameter}
Fig.~\ref{Fig_hyperparam_Appdix} provides results on SleepEDF and FaultDiagnosis datasets for §$4.7$ with consistent observations.

\begin{figure}[t]
\begin{center}
\begin{tabular}{c}
\includegraphics[width = 0.95\linewidth]{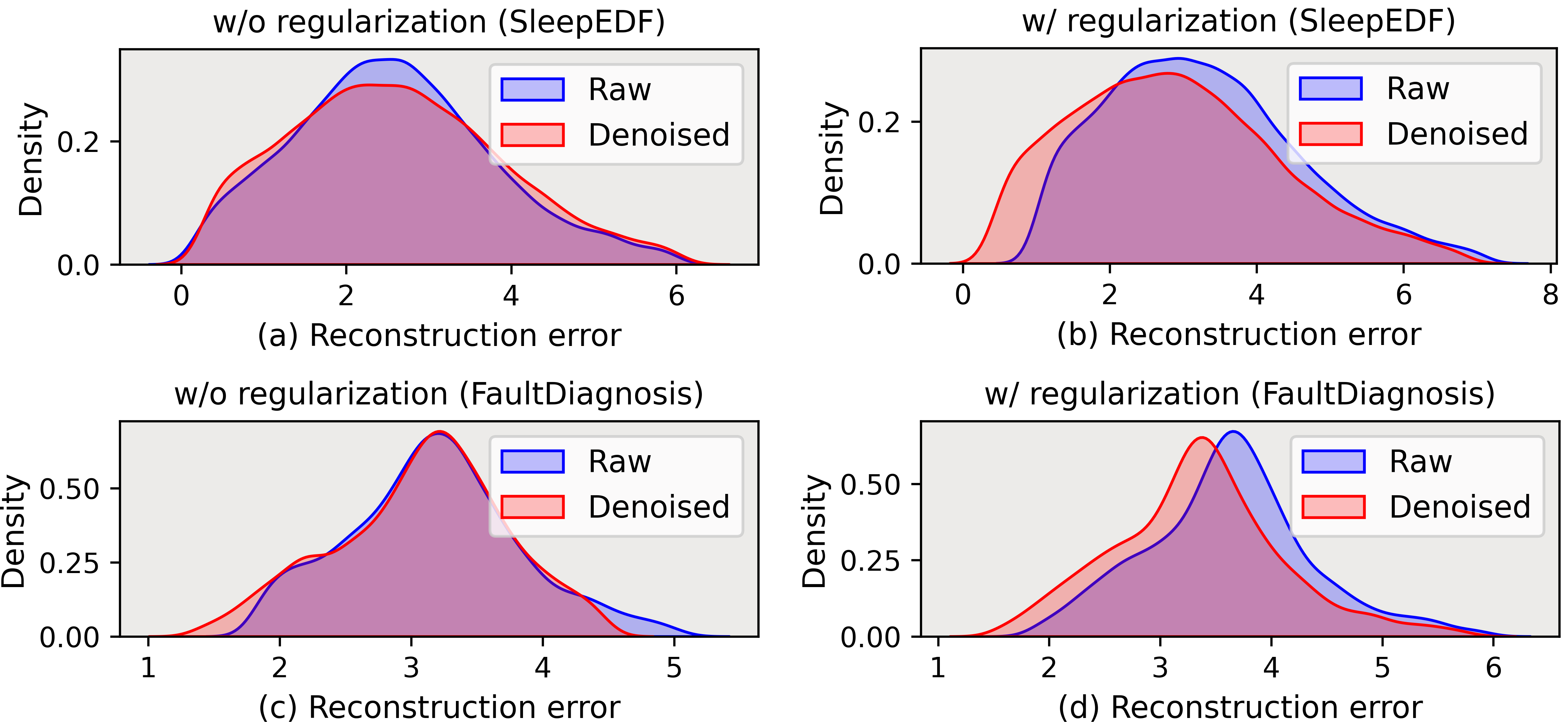}
\end{tabular}
\end{center}
\caption{Reconstruction error analysis on other datasets.}
\label{Fig_reconstrc_error_Appdix}
\end{figure}

\subsection{Visualization of Reconstruction Error Distribution}\label{appendx_exp_reconstruction}
Fig.~\ref{Fig_reconstrc_error_Appdix} provides results on other datasets for §$4.8$ with consistent observations.

\begin{figure*}
\begin{center}
\includegraphics[width = 0.95\linewidth]{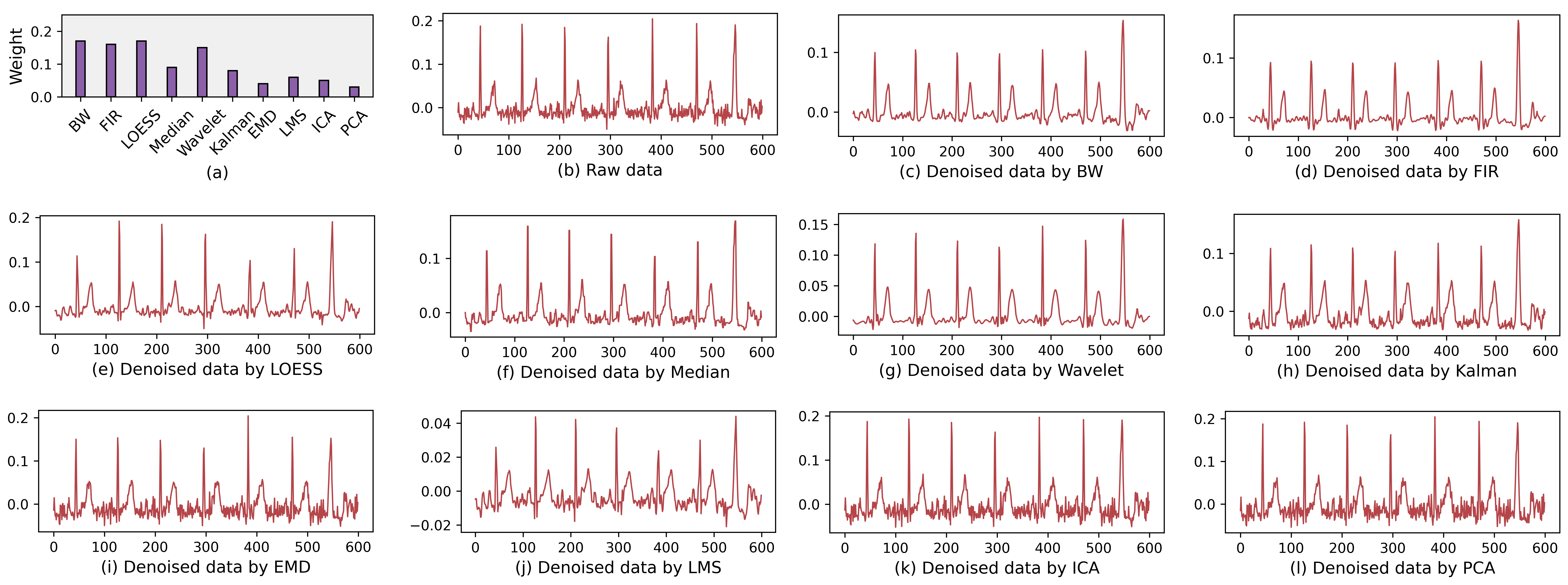}
\end{center}
\caption{A case study about the effect of denoising methods and the assigned weights.
}
\label{Appedx_denosing_effect_wave_plot}
\end{figure*}

\subsection{Analysis of the Selected Denoising Methods}\label{appendx_exp_denoising_methods}
As mentioned in the paper, we investigate whether DECL can select suitable denoising methods and assign them more weights. Specifically, we use a case study on the PTB-XL dataset to (\textit{i}) visualize the weight values, and (\textit{ii}) depict raw data and the denoised counterparts. Notably, we only display one channel of the ECG data and choose a segment of the sample.
A part of the results are shown in §$4.8$ and we present the full results in Fig.~\ref{Appedx_denosing_effect_wave_plot}. In brief, the visualization results support our claim.

\begin{figure}[t]
\begin{center}
\begin{tabular}{c}
\includegraphics[width = 0.95\linewidth]{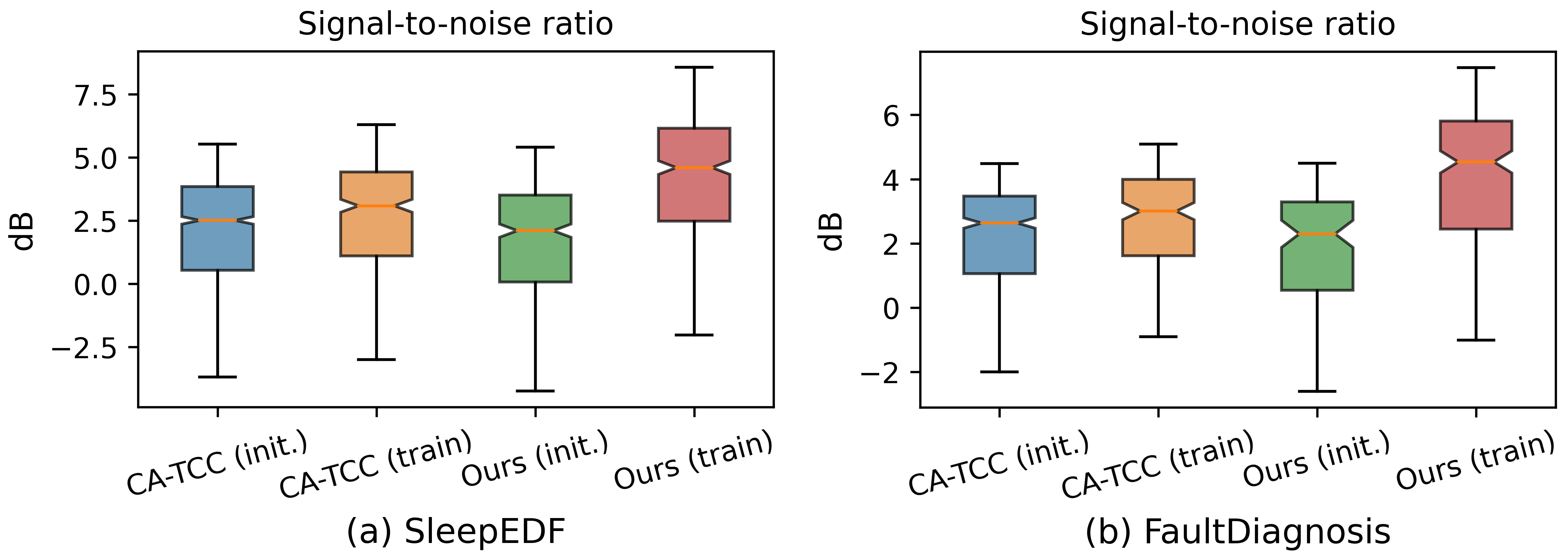}
\end{tabular}
\end{center}
\caption{The comparison of SNR on representation.}
\label{Fig_emb_SNR_box_Appdix}
\end{figure}

\subsection{Analysis on the Learned Representations}\label{appendx_exp_learned_embeddings}
As discussed in §$4.8$, we inject Gaussian noise into the data and then explore whether our method can mitigate noise in representation learning. In detail, we process three datasets (SleepEDF, FaultDiagnosis, and PTB-XL) by setting the std value of the Gaussian distribution as $0.3$. 
The details of the SNR computation can be referred to §\ref{appndx_pilot_study}.
The results of the SleepEDF and FaultDiagnosis dataset are shown in Fig.~\ref{Fig_emb_SNR_box_Appdix}.

\begin{table*}[t]
\centering
\caption{Summary of the collected conventional denoising methods for time series.
}
\resizebox{0.98\linewidth}{!}{
\begin{tabular}{|cc|cc|cc|}
\hline
\multicolumn{2}{|c|}{\textbf{ECG}}                                                                                                                                                        & \multicolumn{2}{c|}{\textbf{EEG}}                                                                                                                                                      & \multicolumn{2}{c|}{\textbf{General   domain}}                                                                                                                                         \\ \hline
\multicolumn{1}{|c|}{\textbf{Denosing   methods}} & \textbf{Hyper-parameters}                                                                                                             & \multicolumn{1}{c|}{\textbf{Denosing methods}} & \textbf{Hyper-parameters}                                                                                                             & \multicolumn{1}{c|}{\textbf{Denosing methods}} & \textbf{Hyper-parameters}                                                                                                             \\ \hline
\multicolumn{1}{|c|}{Butterworth   filter}        & \begin{tabular}[c]{@{}c@{}}order=5,   high freq.=45,\\      low freq.=3\end{tabular}                                                  & \multicolumn{1}{c|}{Butterworth filter}        & \begin{tabular}[c]{@{}c@{}}order=9,   high freq.=40,\\      low freq.=0.3\end{tabular}                                                & \multicolumn{1}{c|}{Butterworth filter}        & \begin{tabular}[c]{@{}c@{}}order=5,   high freq.=45,\\      low freq.=3\end{tabular}                                                  \\ \hline
\multicolumn{1}{|c|}{LOESS}                       & fraction=0.01                                                                                                                         & \multicolumn{1}{c|}{Wiener filter}             & default                                                                                                                               & \multicolumn{1}{c|}{LOESS}                     & fraction=0.01                                                                                                                         \\ \hline
\multicolumn{1}{|c|}{FIR filter}                  & low freq.=3, high   freq.=45                                                                                                          & \multicolumn{1}{c|}{FIR filter}                & low freq.=0.3, high   freq.=40                                                                                                        & \multicolumn{1}{c|}{FIR filter}                & low freq.=3, high   freq.=45                                                                                                          \\ \hline
\multicolumn{1}{|c|}{Median   filter}             & default                                                                                                                               & \multicolumn{1}{c|}{IIR filter}                & f0=40, Q=20                                                                                                                           & \multicolumn{1}{c|}{Median filter}             & default                                                                                                                               \\ \hline
\multicolumn{1}{|c|}{Kalman filter}               & \begin{tabular}[c]{@{}c@{}}transition\_covariance=0.1,\\      observation\_covariance=0.1,\\      transition\_matrices=1\end{tabular} & \multicolumn{1}{c|}{Kalman filter}             & \begin{tabular}[c]{@{}c@{}}transition\_covariance=0.1,\\      observation\_covariance=0.9,\\      transition\_matrices=1\end{tabular} & \multicolumn{1}{c|}{Kalman filter}             & \begin{tabular}[c]{@{}c@{}}transition\_covariance=0.1,\\      observation\_covariance=0.1,\\      transition\_matrices=1\end{tabular} \\ \hline
\multicolumn{1}{|c|}{Wavelet   transform}         & level=3                                                                                                                               & \multicolumn{1}{c|}{Wavelet transform}         & level=5                                                                                                                               & \multicolumn{1}{c|}{Wavelet transform}         & level=3                                                                                                                               \\ \hline
\multicolumn{1}{|c|}{EMD}                         & default                                                                                                                               & \multicolumn{1}{c|}{EMD}                       & default                                                                                                                               & \multicolumn{1}{c|}{Wiener filter}                       & default                                                                                                                               \\ \hline
\multicolumn{1}{|c|}{LMS filter}                  & n=12, mu=0.01                                                                                                                         & \multicolumn{1}{c|}{CCA}                       & n\_components=1000                                                                                                                    & \multicolumn{1}{c|}{LMS filter}                & n=1, mu=0.01                                                                                                                          \\ \hline
\multicolumn{1}{|c|}{ICA}                         & n\_components=1000                                                                                                                    & \multicolumn{1}{c|}{ICA}                       & n\_components=1000                                                                                                                    & \multicolumn{1}{c|}{ICA}                       & n\_components=1000                                                                                                                    \\ \hline
\multicolumn{1}{|c|}{PCA}                         & n\_components=1000                                                                                                                    & \multicolumn{1}{c|}{PCA}                       & n\_components=1000                                                                                                                    & \multicolumn{1}{c|}{PCA}                       & n\_components=1000                                                                                                                    \\ \hline
\end{tabular}
}
\label{Table_denoise_summary}
\end{table*}

\section{Implementation Details}

\subsection{Summary of the Adopted Denoising Methods}\label{appendx_exp_denoise_details}
As mentioned in §$3.3$, we collect a set of commonly used denoising methods $\mathcal{M} = \left\{\phi_1, \phi_2, \ldots, \phi_m\right\}$ from related works. 
Notably, the principle of collecting the denoising methods is to maintain a large diversity and cover various noise types.
This is because denoising methods with varying 
 working mechanisms are adept at handling different noise types.
Take ECG data for example, the finite impulse response (FIR) filter can handle high-frequency noise~\cite{chandrakar2013survey}, the Kalman filter is proven effective for baseline wandering~\cite{mneimneh2006adaptive}, and the wavelet transform can mitigate the muscle artifacts in ECG~\cite{zhang2021heartbeats}.
Following the principle, we respectively collect $10$ methods for different types of time series.
For ECG data and general domain time series, the denoising methods include Butterworth low pass filter~\cite{fang2021dual}, LOESS smoother~\cite{burguera2018fast}, FIR filter~\cite{chandrakar2013survey}, median filter~\cite{vaid2023foundational}, Kalman filter~\cite{mneimneh2006adaptive}, wavelet transform~\cite{zhang2021heartbeats}, empirical mode decomposition (EMD)~\cite{lu2009model},  least mean square (LMS) adaptive filter~\cite{wu2009filtering}, ICA~\cite{chawla2011pca}, and PCA~\cite{romero2011pca}. 
For EEG data, we further leverage the Wiener filter~\cite{kotte2020methods}, infinite impulse response (IIR) filter~\cite{singh2016comparative}, and CCA~\cite{chen2017use}.
An overview of the denoising methods and their hyper-parameters are summarized in Table~\ref{Table_denoise_summary}.

\subsection{Implementation Details of DECL}\label{appendx_exp_DECL_details}
Here, we introduce the implementation details of the encoder and auto-regressive (AR) module in the proposed method DECL.
The encoder is a $3$-block convolutional architecture. 
Each block consists of a $1$D-convolutional layer, a batch normalization layer, an activation layer, a max-pooling layer, and one dropout layer. Furthermore, the convolutional layers have a filter width of $8$ and contain $32\ast2\textit{t}$ filters, with $\textit{t}$ being a value that starts from $1$ and increases by $1$ for every block.
Its final output channel number $r$ is $128$.
The dimension size of the presentation $C$ for the five datasets is respectively $127$, $642$, $44$, $44$, and $44$.
As for the AR module, we implement it with a Transformer~\cite{vaswani2017attention} of $L$ successive blocks of multi-headed attention layer and MLP block.
Specifically, the MLP block consists of two fully connected layers with a rectified linear activation unit (ReLU) and a dropout layer in between. 
Here, we set the number of identical blocks $L$ as $4$, assign the number of heads as $4$, and set its hidden dimension size $h$ as $100$.

\subsection{Implementation Details of Pilot Study}\label{appendx_exp_pilot_details}
\label{appndx_pilot_study}
We provide details for the motivating analysis in the \textit{Introduction} section.
We respectively apply two denoising methods (i.e., LOESS and median filter) for the PTB-XL dataset in the pre-processing and then examine the performance of two SSL methods (i.e., CA-TCC and TS2vec) on the denoised data.
We perform linear evaluation to obtain the downstream classification results and the setup is illustrated in the §$4.3$.
In addition, we also compare the data noise in the raw data and the representations of the two models. 
We adopt signal-to-noise ratio (SNR) which is a widely used metric to quantize the data noise in time series~\cite{dera2023trustworthy}. 
A higher SNR value indicates that the time series contains less noise.
Since the information about data noise is generally unavailable in the datasets, we induce Gaussian noise into the data by setting the std value as $0.5$. 
Given a time series sample $\boldsymbol{x}_i =\left[x_{i, 1}, x_{i, 2}, \ldots, x_{i, T}\right]$, we use $\mathbf{\bar{s}}_i =\left[\bar{s}_{i, 1}, \bar{s}_{i, 2}, \ldots,  \bar{s}_{i, T}\right]$ to denote the injected noise for $\boldsymbol{x}_i$ and use $\boldsymbol{\bar{x}}_i =\left[\bar{x}_{i, 1}, \bar{x}_{i, 2}, \ldots, \bar{x}_{i, T}\right]$ to present the noise-induced counterpart.
Accordingly, the SNR value of $\boldsymbol{\bar{x}}_i$ can be calculated via the following formula:
\begin{equation}
\begin{gathered}
\text{SNR}_\text{noisy} = 10 \times \lg \left(\frac{\sum_{j=1}^T \bar{x}_{i, j}^2}{\sum_{j=1}^T \bar{s}_{i, j}^2}\right).
\end{gathered}
\label{Eq_SNR_add_noise}
\end{equation}
Then, we employ LOESS for noise removal on the dataset and compute the SNR value of the denoised data $\check{\boldsymbol{x}}_{i}$ as:
\begin{equation}
\begin{gathered}
\text{SNR}_\text{denoised} = 10 \times \lg \left(\frac{\sum_{j=1}^T  \check{x}_{i, j}^2}{\sum_{j=1}^T \left(\check{x}_{i, j} - x_{i, j} \right)^2}\right),
\end{gathered}
\label{Eq_SNR_add_noise_denoise}
\end{equation}
where $\check{x}_{i, j}$ is a timestep of $\check{\boldsymbol{x}}_{i}$.
Later, we map the denoised data into latent space.
Let $\mathbf{z}_{i} =\left[z_{i, 1}, z_{i, 2}, \ldots,  z_{i, C}\right]$ and $\check{\mathbf{z}}_{i} =\left[\check{z}_{i, 1}, \check{z}_{i, 2}, \ldots,  \check{z}_{i, C}\right]$ denote the representation of the orignal data $\boldsymbol{x}_{i}$ and the denoised data $\check{\boldsymbol{x}}_{i}$, respectively. 
Following related works~\cite{Vu2017DataEI}, the SNR formula of the representation is written as:
\begin{equation}
\begin{gathered}
\text{SNR}_\text{rep.} = 10 \times \lg \left(\frac{\sum_{j=1}^C  \check{z}_{i, j}^2}{\sum_{j=1}^C \left(\check{z}_{i, j} - z_{i, j} \right)^2}\right),
\end{gathered}
\label{Eq_SNR_add_noise_denoise_rep}
\end{equation}
where $C$ is the dimension size of the representation.

\section{Related Works}

\subsection{Time Series Self-supervised Learning}\label{appendx_exp_realted_works}
We further introduce existing time series SSL methods from three aspects: contrastive-based, generative-based, and adversarial-based~\cite{zhang2023self_SSL}\cite{meng2023unsupervised}.
The contrastive-based approaches rely on different assumptions and the manners of building similar (positive) and dissimilar (negative) samples~\cite{chen2020simple} are varied. Some representative methods are as follows. 
(\textit{i}) Prediction-based methods~\cite{eldele2021time}\cite{schneider2022detecting} are featured by using historical data to predict future information, in which informative features can be learned. For instance, CPC~\cite{oord2018representation} adopts the data in the future timesteps as positive samples and randomly samples data from different batches as negative samples for contrastive learning.
(\textit{ii}) Augmentation-based methods~\cite{kiyasseh2021clocs}\cite{yang2022timeclr} apply various data augmentation methods to preserve the feature consistency of different data views. For example, BTSF~\cite{yang2022unsupervised_icml} exploits the temporal-spectral affinities of time series for building contrastive pairs.
(\textit{iii}) Sampling-based methods~\cite{yeche2021neighborhood}\cite{tonekaboni2021unsupervised} assume that neighboring regions of time series maintain a high degree of similarity and thereby directly sample different segments of the raw data for contrastive learning.
(\textit{iv}) Prototype-based methods~\cite{MengQLCX023} rely on cluster assumption (samples of the same class tend to form a cluster) and incorporate prototypes (cluster centers) in contrastive learning. 
(\textit{v}) Domain knowledge-based methods~\cite{nonnenmacher2022utilizing} further leverage expert prior knowledge to build suitable positive and negative samples to guide model training. 
Differing from the contrastive learning methods, generated-based methods mainly rely on reconstruction for representation learning~\cite{wang2022learning}\cite{li2020learning}\cite{vaid2023foundational}. 
Adversarial-based methods usually adopt a generator and discriminator for adversarial learning~\cite{jeon2022gt}\cite{jeha2021psa}\cite{luo2019e2gan}.
However, the above methods generally assume time series is free of noise and may not well mitigate noise in representation learning.

\end{document}